\documentclass[preprint,12pt]{elsarticle}
\usepackage[colorlinks=true,linkcolor=blue,citecolor=blue,urlcolor=blue,breaklinks=true]{hyperref}
\usepackage{xurl}
\usepackage{tabularx}
\usepackage{subcaption}
\usepackage{longtable}
\usepackage{multirow}
\usepackage{makecell}
\usepackage{amsmath}
\usepackage{amssymb}
\usepackage{float}
\usepackage{comment}
\usepackage{natbib}
\usepackage{algorithm}
\usepackage{algorithmic}
\usepackage{booktabs}
\usepackage{threeparttable}
\usepackage{rotating}

\graphicspath{{figures/}{figures/noor_20/}{figures/noor_5/}{figures/c8_1/}{figures/c8_10/}}

\newcommand{\indic}{\mathbf{1}}

\journal{Results in Engineering}

\begin{document}

\begin{frontmatter}

\title{TRIAGE: Risk-Controlled Pseudo-Label Admission for Annotation-Efficient Semi-Supervised Retinal OCT Classification}

\author[1]{Md Ashraful Hossen Akash}
\author[1]{Shyla Afroge}
\author[1]{Abdullah Al Mamun}
\author[2]{Md. Kishor Morol}
\author[3,4]{\texorpdfstring{Tze Hui Liew\corref{cor1}}{Tze Hui Liew}}
\ead{thliew@mmu.edu.my}

\cortext[cor1]{Corresponding author.}

\address[1]{Department of Computer Science \& Engineering, Rajshahi University of Engineering \& Technology, Rajshahi-6204, Bangladesh}
\address[2]{Elite Research Lab LLC, New York, USA}
\address[3]{Faculty of Information Science and Technology, Multimedia University, Melaka, Malaysia}
\address[4]{Centre for Intelligent Cloud Computing (CICC), COE of Advanced Cloud, Faculty of Information Science \& Technology, Multimedia University, 75450, Melaka, Malaysia}

\begin{abstract}

The advanced retinal disease diagnosing imaging modality, optical coherence tomography (OCT), encounters a lack of automation because of the high expenses for annotations performed by specialists. The use of SSL solves the problem of insufficient annotations using unlabeled B-scans; however, most of the current techniques for generating pseudo-labels are based on prediction confidence without considering the asymmetry between different types of errors. This paper proposes TRIAGE, a risk-controlled semi-supervised framework for OCT scans classification, which uses the concept of a patient-level conformal risk controller with an asymmetric cost matrix. TRIAGE unites three crucial modules: a hierarchical classifier that is capable of working with partially abnormal supervision of the disease subtypes, a patient-grouped conformal risk controller with primal-dual coverage control, and a context-aware Transformer teacher for cross-slice verification. On the dataset from Noor Eye Hospital (16,822 B-scans, 161 patients, and 554 volumes) with a test set of unseen patients, TRIAGE demonstrates 89.66\% scan-level accuracy, 0.8805 macro-F1, 0.9641 macro-AUC, and an 8.34\% under-grading rate when using only 20\% of the labeled data. With only 5\% of the labeled data, TRIAGE keeps 76.88\% accuracy and a 0.1656 under-grading rate. Compared with the other six state-of-the-art semi-supervised methods, TRIAGE significantly outperforms them with ablation study demonstrating the contribution of each module in the overall framework performance (by 42.7\% in terms of under-grading rate comparing to fixed threshold methods). TRIAGE demonstrates 98.00\% accuracy for 3-class classification with 1\% labeled data and 95.94\% accuracy for 8-class classification with 10\% labeled data on the OCT-C8 dataset.

\end{abstract}


\end{frontmatter}

% =====================================================================
\section{Introduction}
\label{sec:intro}

Optical coherence tomography (OCT) is considered the gold standard imag-
ing technique to identify vision threatening retinal diseases such as Age Related
Macular Degeneration (AMD) and Diabetic Macular Edema (DME)~\citep{huang1991oct,aumann2019oct}.
Automated classification of OCT B-scan images helps accelerate clinical triage and
solve specialist reading bottleneck problems~\citep{kermany2018identifying,sulaiman2025review}. However, training deep networks needs tens of thousands of OCT B-scans annotated by medical experts
at an extraordinarily high cost. Semi-Supervised Learning (SSL) mitigates annotation problems by using unlabelled data via pseudo-labelling or consistency
regularization~\citep{jiao2024limited,solatidehkordi2022survey,singh2024machine,cheplygina2019}. Conventional semi-supervised learning methods like Pseudo-Label~\citep{lee2013pseudo}, Mean Teacher~\citep{tarvainen2017mean},  FixMatch~\citep{sohn2020fixmatch}, CoMatch~\citep{li2021comatch}, FlexMatch~\citep{zhang2021flexmatch}, and FreeMatch~\citep{wang2023freematch} use a fixed or adaptive threshold to accept pseudo-labels.

However, thresholding based on confidence suffers from two critical limitations in
clinical deployment:

\begin{enumerate}
    \item \textbf{Asymmetric Clinical Risk}: Conventional confidence thresholds treat all types of errors as equivalent. Classifying a choroidal neovascularisation (CNV) scan as \textsc{normal} would delay referrals dangerously whereas classifying drusen as CNV would lead to only over-referrals. Confidence thresholding does not consider this asymmetric risk structure at all.
    
    \item \textbf{Confirmation Bias \& Decoupling}: In extreme conditions of label
    scarcity, deep neural networks become too over-confident about incorrect predictions resulting in erroneous pseudo-labels corrupting student network training~\citep{yangHABIT,hangRSSL}.

    \item \textbf{Unused Spatial Volume Context}: B-scans are sequential slices in retinal 3D volumes. Current OCT SSL approaches ignore important spatial context information between B-scans treating them independently.
\end{enumerate}

To tackle such issues, we propose \textbf{TRIAGE} – a clinical-risk controlled semi-supervised OCT classifier. Rather than arbitrary thresholding, TRIAGE controls pseudo-label admissions using distribution-free clinical risk bounds calibrated with patient-grouped conformal risk control~\citep{angelopoulos2024conformal}.

TRIAGE integrates three core engineering innovations:
\begin{itemize}
    \item \textbf{Hierarchical Classifier Backbone}: The decision space is divided into two heads – one coarse-grained (\textsc{normal}/\textsc{abnormal}) and one fine-grained (\textsc{drusen}/\textsc{cnv}). For scans with clear abnormality and unclear subtypes, partially supervised abnormal labels are assigned.

    \item \textbf{Patient-Grouped Conformal Risk Controller}: Admissions threshold $\lambda$ is calibrated against asymmetric clinical cost matrix with Hoeffding–Bentkus upper confidence bounds on a patient-disjoint calibration data. Primal-Dual Coverage control actively avoids under-admission during early training.
    
    \item \textbf{Volume-Aware Context Teacher}: Lightweight Transformer encoder is used to enforce the cross-B-scans spatial agreement in 3D volumes between the context teacher and the single scan teacher prior to admissions~\citep{vaswani2017attention}.
\end{itemize}

TRIAGE achieves 89.66\% accuracy and 8.34\% under-grading at 20\% labels, as well as 76.88\% accuracy at 5\% labels in a strict patient-disjoint setting on the Noor Eye Hospital dataset (16,822 B-scans, 161 patients, 554 volumes), outperforming six SSL baselines. On the OCT-C8 dataset, TRIAGE achieves 98.00\% accuracy for 3-class case at 1\% labels and 95.94\% for 8-class case at 10\% labels.

The paper is organized as follows: Section~\ref{sec:related} surveys related work; Section~\ref{sec:method} describes TRIAGE; Section~\ref{sec:setup} describes experiments; Section~\ref{sec:results} discusses results; Section~\ref{sec:discussion} discusses implications for medicine; Section~\ref{sec:conclusion} concludes.

% =====================================================================
\section{Related Work}
\label{sec:related}

\subsection{Supervised \& Architectural Advancements in Retinal Image Analysis}

Deep CNN interpretation of OCT was first introduced by Kermany
et al.~\citep{kermany2018identifying} that showed CNN could perform comparably to human medical specialists in detecting retinal diseases like age-related macular degeneration (AMD), and diabetic macular edema (DME). Further extending from these
works, Gu et al.~\citep{guFundus} applied deep learning to grade multi-severity diabetic retinopathy, and Song et al.~\citep{songDRT} suggested the application of a deep relationship transformer for joint analysis of visual field defects and OCT structural changes. In a very recent work published in \textit{Results in Engineering}, Mohanraj, Karthika et al.~\citep{mohanraj2026advancements} used the pathology-driven pre-processing and multi-model CNN ensemble for emphasizing vascular bulges and fluid features of retina. In a recent review paper, Sulaiman et al.~\citep{sulaiman2025review} stressed that, although supervised classifiers provide very high diagnosis accuracy, they still have performance limitations related to the quality of expert annotations.

In order to address issues such as speckle noise and spatial scale variations prevalent in OCT B-scans, research into advanced convolution operations has been carried out. In particular, Peng et al.~\citep{pengMSDRCN} proposed a multi-scale denoising residual network that removes speckle noise without smearing fine boundaries of retinal layers, whereas Li et al.~\citep{liPreprocess} defined structure-oriented preprocessing strategies to achieve normalization of intensity distributions. Lightweight models have been developed for use in constrained clinical environments; for instance, Pan et al.~\citep{panLightweight} proposed a small CNN with a 60\% reduction in model size, Subramanian et al.~\citep{subramanian2022classification} studied deep transfer learning with CNNs for OCT screening, Karthik and Mahadevappa~\citep{karthikAdaptive} proposed adaptive convolution blocks for retinal feature extraction, and Ahlawat et al.~\citep{ahlawatVGG} studied modified VGG networks. Recently, Vision Transformers have been explored to capture long-range dependencies between retinal layers. Shen et al.~\citep{shenTransformer} proposed a structure-oriented transformer for disease severity grading, He et al.~\citep{heTransformer} used interpretable cross-attention maps for explaining OCT classification, and Malik et al.~\citep{malikAPSID} used an attention-prototype framework (APSID) for multi-scale feature learning. However, these supervised approaches require large-scale annotated datasets and lack risk-aware gating to avoid under-grading.

\subsection{Semi-Supervised Learning \& Generative Frameworks in Ophthalmic Imaging}

Semi-supervised learning (SSL) alleviates the problem of annotation shortages
through consistency regularization, pseudo-labeling, and generative modeling~\citep{jiao2024limited,solatidehkordi2022survey,singh2024machine,cheplygina2019}. Adversarial semi-supervised learning has proven successful in fea-
ture extraction tasks in industries and medicine. For example, Luo et al.~\citep{luo2026semisupervised} introduced the SSGD-Seg model in \textit{Results in Engineering} that shows how
dynamic convolution and discriminator guided adversarial training can lead
to an accurate detection of defects using only 20\% of the data labeled.

In the field of ophthalmic image analysis, semi-supervised learning approaches have been adapted for domain shifts and annotation limitations. The out-of-distribution glaucoma detection was considered by Wang et al.~\citep{wangGlaucomaOOD} and Narasimha et al.~\citep{narasimhaGlaucoma} who used uncertainty gating and multi-agent deliberation for reliable filtering of predictions. Huang et al.~\citep{huangTLSDA} designed subdomain adaptation for OCT retinopathy, Qiu and Zheng~\citep{qiuDualEncoder} introduced the dual-encoder attention network for unlabelled B-scan classification, while Li et al.~\citep{liCEP} created complementary expert pooling for aggregation of unla-
beled feature streams. The work most similar to ours was done by Li et al.~\citep{liRetinopathy} who used contrastive feature learning combined with FixMatch for OCT classification and Alizadeh et al.~\citep{jodeiriITS} who performed teacher-student consistency experiment on the Noor dataset. However, existing approaches to opththalmic semi-supervised learning use scalar confidence thresholding that is not sensitive to asymmetric clinical costs, partial subtype ambiguities and 3D volume continuity.

\subsection{Reliable Pseudo-Labelling \& Confirmation Bias Mitigation}

The core problem in the domain of pseudo-labelling during semi-supervised learning is confirmation bias where the student model makes overconfident predictions of incorrect labels for unlabelled images leading to corruption in gradients~\citep{yangHABIT,hangRSSL}. Self-paced sampling by Guan et al.~\citep{guanS2Match} and similarity-based prototype alignment for constraining pseudo-label drifts by Mahmood et al.~\citep{mahmoodSPLAL} are some of the efforts towards improving pseudo-label quality. Adaptive confidence thresholding strategies
are another popular way of improving pseudo label quality and include classadaptive thresholding by Yang et al.~\citep{yangCLCP}, confidence-calibrated contrastive mean teachers by Wang et al.~\citep{wangC3MT}, and combination of multi-scale feature consistency and adversarial learning by Shiyan et al.~\citep{shiyanFCAT}. More recent efforts in prompt-driven and graph regularized SSL include PromptMed by Wang and Mao~\citep{wangPromptMed} to provide consistency across class distribution, reciprocal collaboration of classification heads by Zeng et al.~\citep{zengReco}, ADGNET for reducing medical annotations by Yang~\citep{yangADGNET}, and distance correlation minimization by Berenguer et al.~\citep{berenguerDistance}. All of the above methods dynamically adapt their confidence thresholding schemes according to empirical class distribution or distances but do not take into account the clinical risk associated with a misclassified image candidate. Unlike the above schemes, TRIAGE evaluates pseudo label candidates based on explicit clinically motivated risk bound calculations.

\subsection{Multi-Teacher Consistency \& Identified Research Gap}
Multi-teacher, dual-decoder, and mutual teaching strategies enhance pseudo-label quality by enforcing agreement between complementary feature representations. Zhong et al.~\citep{zhongCrossDistill} introduced cross-distillation with CAM consistency, Zhang et al.~\citep{zhangDualDecoder} proposed dual-decoder mutual teaching, and Min et al.~\citep{minCoupling} developed deep non-consistent mean teachers to mitigate model coupling. In medical image segmentation, Wu et al.~\citep{wuDuCiSC} formulated dual cross-image semantic consistency, Zeng et al.~\citep{zengUnCo} designed an uncertainty co-estimator, and Wei et al.~\citep{weiSegMatch} established SegMatch for surgical instrument segmentation. However, these dual-teacher architectures treat individual 2D B-scans as independent samples, ignoring the inherent 3D spatial continuity across adjacent B-scan slices within retinal volumes.

Synthesizing this literature reveals a three-fold research gap in retinal OCT semi-supervised learning:
\begin{enumerate}
    \item \textbf{Absence of Asymmetric Risk Control}: Prior SSL algorithms utilize symmetric confidence thresholds that treat under-grading a diseased scan as \textsc{normal} identically to minor over-referral.
    \item \textbf{All-or-Nothing Pseudo-Label Admission}: Existing methods either admit a hard multi-class label or discard the sample entirely, failing to provide partial supervision when binary abnormality is clear but fine-grained subtyping is ambiguous.
    \item \textbf{Neglect of 3D Spatial Volume Context}: Standard SSL models evaluate 2D B-scans in isolation without enforcing cross-slice spatial verification across volume neighbours.
\end{enumerate}
TRIAGE directly addresses this gap by combining patient-grouped conformal risk calibration, hierarchical partial supervision, and volume-aware Transformer teacher verification within a unified framework.

% =====================================================================
\section{The TRIAGE Framework}
\label{sec:method}

\subsection{Problem Formulation \& Pipeline Overview}
\begin{figure}[htbp]
    \centering
    \includegraphics[width=0.85\textwidth]{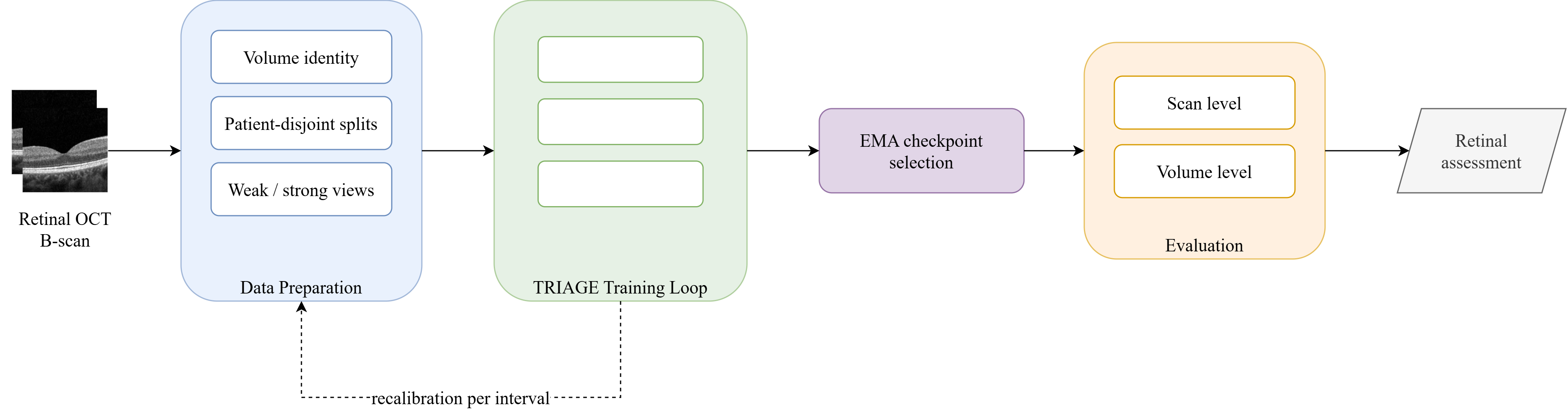}
    \caption{Overall abstract architecture and data pipeline of TRIAGE, illustrating patient-disjoint splitting, weak/strong view generation, training loop, and dual-level evaluation.}
    \label{fig:pipeline}
\end{figure}

Let $\mathcal{D}_l = \{(x_i, y_i)\}_{i=1}^{n_l}$ denote a labelled set of B-scans with diagnostic labels $y_i \in \{1,\dots,K\}$, and $\mathcal{D}_u = \{x_j\}_{j=1}^{n_u}$ a much larger unlabelled pool ($n_u \gg n_l$). Standard semi-supervised learning (SSL) admits unlabelled samples into training whenever the model's top predicted probability exceeds a scalar confidence threshold $\tau$ ($\max_c p(c \mid x) > \tau$). TRIAGE replaces arbitrary confidence thresholding with a clinically governed admission problem: admit candidate pseudo-labels only when their expected clinical risk is certified, at a distribution-free confidence level $1-\delta$, not to exceed a stated risk budget $\alpha$. Figure~\ref{fig:pipeline} presents the overall abstract data pipeline of TRIAGE.

\begin{figure}[htbp]
    \centering
    \includegraphics[width=0.90\textwidth]{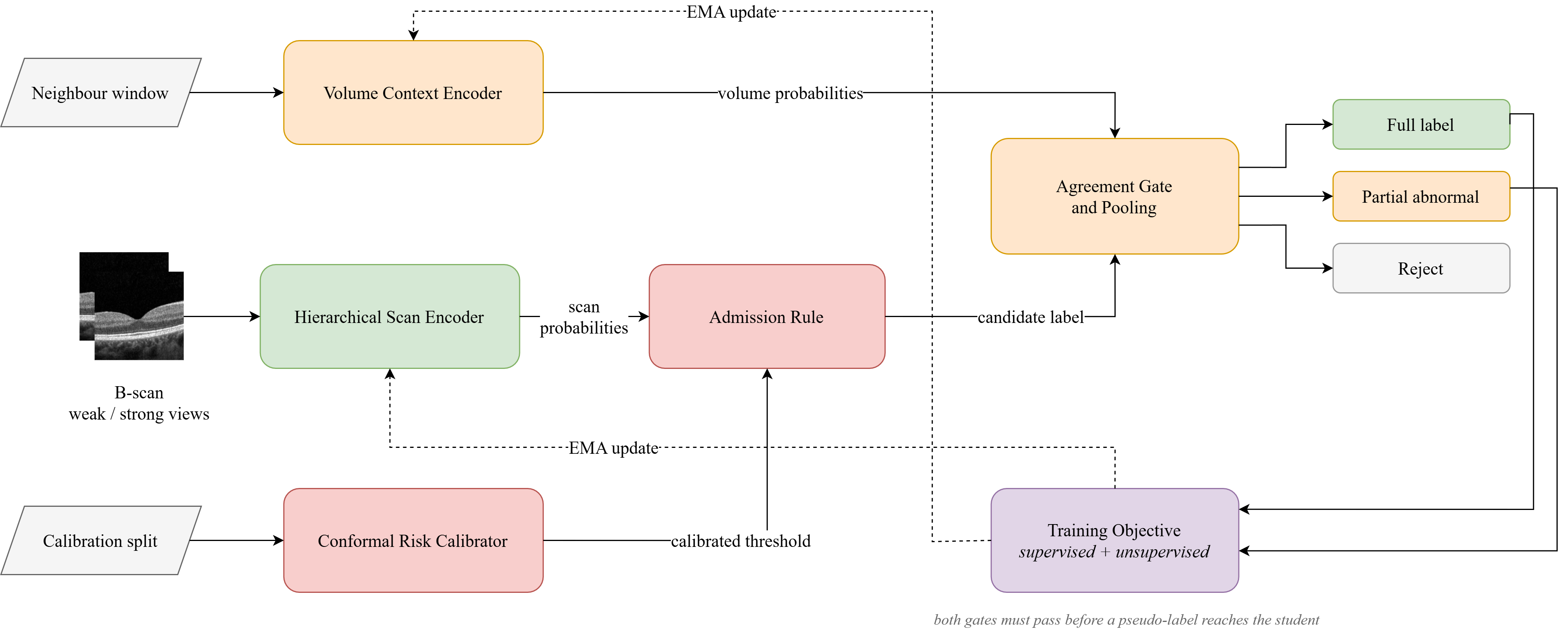}
    \caption{Detailed main model architecture and dual-teacher workflow of TRIAGE, illustrating single-scan risk gating, volume-context Transformer teacher agreement, and student supervision.}
    \label{fig:arch}
\end{figure}

As illustrated in Figure~\ref{fig:arch}, one training iteration of TRIAGE proceeds systematically through seven steps:
\begin{enumerate}
    \item \textbf{Labelled Pass}: Sample a labelled batch $\mathcal{D}_l$ and compute the supervised hierarchical cross-entropy loss over both binary and fine classification heads (Section~\ref{sec:backbone}).
    \item \textbf{Unlabelled Sampling \& Augmentation}: Sample an unlabelled batch $\mathcal{D}_u$. For each target B-scan, retrieve $W=5$ adjacent B-scans from its reconstructed 3D volume key ($\text{Patient ID} + \text{Eye} + \text{Session}$). Generate weak (flip-only) and strong (RandAugment + Cutout) views for the target scan, and weak views for volume neighbours.
    \item \textbf{Dual-Teacher Probability Estimation}: Pass weak views to the single-scan EMA teacher to obtain single-scan hierarchical probabilities, while passing the neighbourhood window to the volume-context EMA teacher to obtain cross-slice verified probabilities (Section~\ref{sec:vista}).
    \item \textbf{Clinical Risk Evaluation}: Calculate expected clinical risk $r(c \mid x)$ across candidate classes using asymmetric cost matrix $R$; apply the patient-calibrated conformal admission rule to determine the candidate label (Full, Partial, or Reject) (Section~\ref{sec:crc}).
    \item \textbf{Dual-Teacher Agreement Gate}: Discard candidate admissions unless the single-scan teacher and volume-context teacher agree at both binary and fine decision levels.
    \item \textbf{Student Optimization}: Train the student model via SGD on the strongly augmented unlabelled target views using the surviving hard or partial pseudo-label supervision. Update single-scan and volume teachers via exponential moving average (EMA decay 0.999).
    \item \textbf{Interval Recalibration}: Recalibrate admission threshold $\lambda$ on the patient-disjoint calibration split using Hoeffding--Bentkus upper confidence bounds, and update the primal--dual coverage multiplier $\mu$.
\end{enumerate}

\subsection{Hierarchical Classifier Backbone \& MBConv Micro-Architecture}
\label{sec:backbone}

\begin{figure}[htbp]
    \centering
    \begin{subfigure}[b]{0.85\textwidth}
        \centering
        \includegraphics[width=\textwidth]{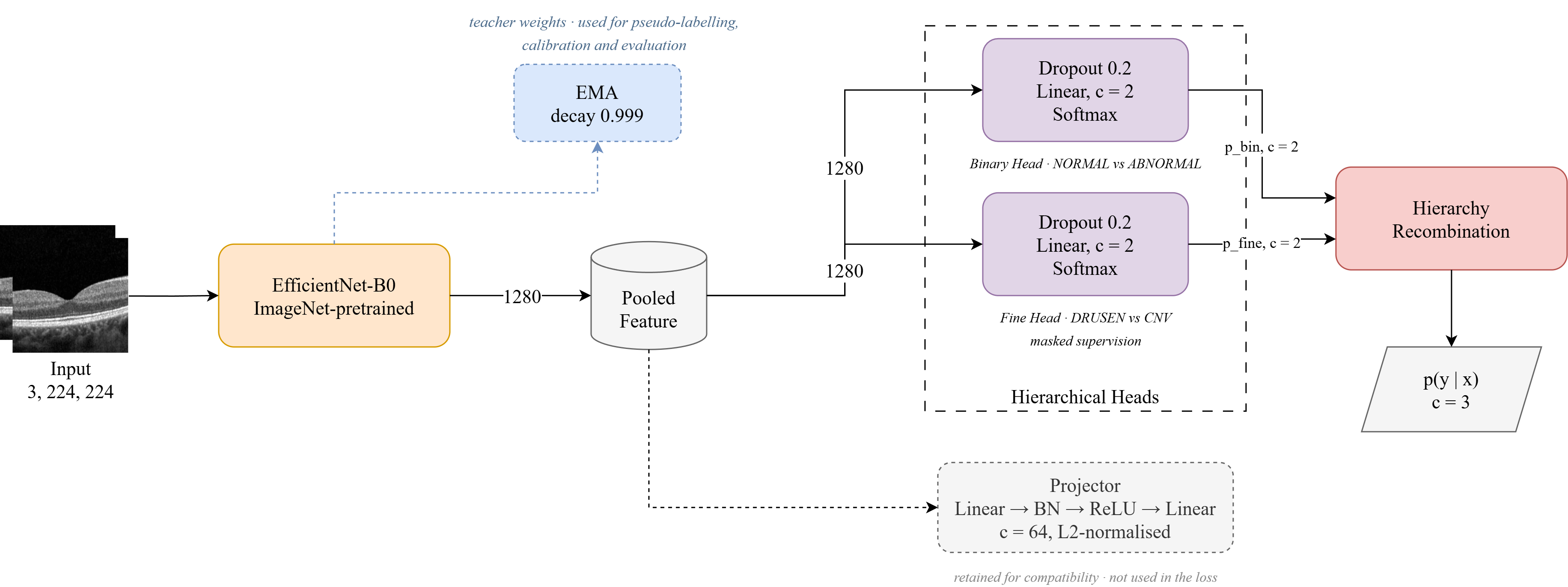}
        \caption{Hierarchical scan encoder and multi-head probability recombination.}
        \label{fig:scan_encoder}
    \end{subfigure}
    \vspace{0.6em}
    \begin{subfigure}[b]{0.85\textwidth}
        \centering
        \includegraphics[width=\textwidth]{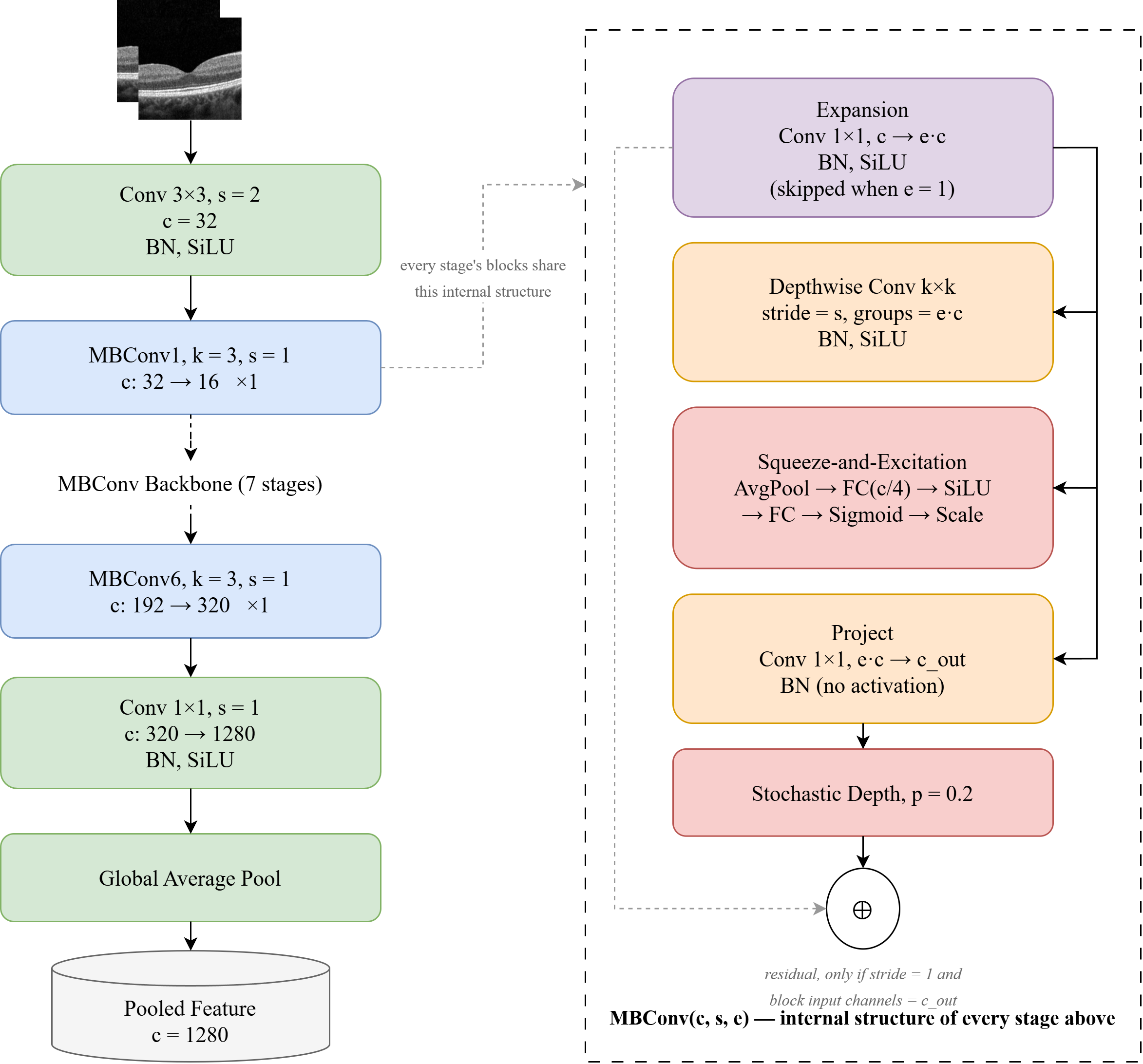}
        \caption{EfficientNet-B0 backbone stages and internal MBConv block structure.}
        \label{fig:heads}
    \end{subfigure}
    \caption{Hierarchical backbone: (a) Feature extraction pipeline with dual classification heads and CoMatch projection head, and (b) 7-stage EfficientNet-B0 backbone with internal MBConv block operations including Squeeze-and-Excitation attention and Stochastic Depth.}
    \label{fig:backbone_layout}
\end{figure}

The single-scan feature extractor uses an ImageNet-pretrained EfficientNet-B0 backbone~\citep{tan2019efficientnet}, selected for its optimal trade-off between representation capacity and inference FLOPs. Figure~\ref{fig:backbone_layout} illustrates the feature extraction pipeline, dual classification heads, and the detailed internal structure of its Mobile Inverted Bottleneck Convolution (MBConv) blocks.

\subsubsection{Backbone \& MBConv Block Micro-Architecture}
As detailed in Figure~\ref{fig:backbone_layout}(b), an input B-scan $x \in \mathbb{R}^{3 \times 224 \times 224}$ is first processed by a stem Conv $3 \times 3$ (stride 2, 32 channels, BatchNorm, SiLU activation) followed by 7 stacked MBConv stages. Each MBConv block incorporates four sequential operations:
\begin{enumerate}
    \item \textbf{Channel Expansion}: A $1 \times 1$ point-wise convolution expands input channels $c$ by an expansion factor $e$ ($c \rightarrow e \cdot c$), followed by BatchNorm and SiLU activation (skipped when $e=1$).
    \item \textbf{Depthwise Convolution}: A $k \times k$ depthwise separable convolution (kernel $k \in \{3,5\}$, stride $s \in \{1,2\}$) applies spatial filtering independently across $e \cdot c$ groups.
    \item \textbf{Squeeze-and-Excitation (SE) Channel Attention}: Global average pooling compresses spatial dimensions into a channel descriptor $\mathbf{z} \in \mathbb{R}^{e \cdot c}$, which passes through a two-layer FC bottleneck (reduction ratio 4, SiLU activation, Sigmoid gating) to compute dynamic channel recalibration weights scaling feature channels.
    \item \textbf{Linear Projection \& Residual Connection}: A $1 \times 1$ point-wise convolution projects channels back to $c_{\text{out}}$ without non-linear activation. When stride $s=1$ and $c = c_{\text{out}}$, a residual skip connection with Stochastic Depth regularisation ($p=0.2$) adds the input feature directly to the block output.
\end{enumerate}
A final $1 \times 1$ Conv layer expands features to 1,280 channels before Global Average Pooling yields the 1,280-dimensional feature vector $\mathbf{f} \in \mathbb{R}^{1280}$.

\subsubsection{Hierarchical Heads \& Probability Recombination}
Rather than deploying a flat 3-class softmax over $\mathbf{f}$, TRIAGE factors the decision space into two decoupled linear heads (Figure~\ref{fig:backbone_layout}(a)):
\begin{itemize}
    \item \textbf{Binary Head}: A linear layer $\mathbb{R}^{1280} \rightarrow \mathbb{R}^2$ with Dropout ($p=0.2$) predicts binary disease status $p_{\text{bin}} = [p_{\text{bin}}(\textsc{normal}), p_{\text{bin}}(\textsc{abnormal})]$.
    \item \textbf{Fine Head}: A linear layer $\mathbb{R}^{1280} \rightarrow \mathbb{R}^2$ with Dropout ($p=0.2$) predicts fine-grained disease subtype $p_{\text{fine}} = [p_{\text{fine}}(\textsc{drusen}), p_{\text{fine}}(\textsc{cnv})]$, trained exclusively on abnormal samples via masked cross-entropy.
    \item \textbf{Projection Head}: A 2-layer MLP (Linear $\rightarrow$ BN $\rightarrow$ ReLU $\rightarrow$ Linear) projects $\mathbf{f}$ into an L2-normalised 128-dimensional representation $\mathbf{z} \in \mathbb{R}^{128}$ for compatibility with contrastive graph memory structures.
\end{itemize}

The flat 3-class posterior distribution $p(y \mid x)$ is recovered exactly via chain rule:
\begin{align}
p(\textsc{normal}) &= p_{\text{bin}}(\textsc{normal}), \label{eq:h1}\\
p(\textsc{drusen}) &= p_{\text{bin}}(\textsc{abnormal}) \cdot p_{\text{fine}}(\textsc{drusen}), \label{eq:h2}\\
p(\textsc{cnv}) &= p_{\text{bin}}(\textsc{abnormal}) \cdot p_{\text{fine}}(\textsc{cnv}). \label{eq:h3}
\end{align}
Decoupling binary abnormality detection from fine-grained subtyping allows uncertain disease subtypes to contribute partial \textsc{abnormal} supervision without forcing uncalibrated subtype guesses.

\subsection{Clinical-Risk-Aware Conformal Admission}
\label{sec:crc}

\subsubsection{Asymmetric Clinical Cost}
Retinal screening errors carry asymmetric clinical costs: under-grading a CNV scan as \textsc{normal} delays urgent anti-VEGF referral, whereas confusing drusen with CNV results in manageable referral. We encode this policy in cost matrix $R[y,c]$ (Table~\ref{tab:riskmatrix}). Expected risk for candidate class $c$ is:
\begin{equation}
r(c \mid x) = \sum_{y} R[y,c] \, p(y \mid x).
\label{eq:risk}
\end{equation}

\begin{table}[H]
\centering
\caption{Clinical cost matrix $R[\text{true}][\text{predicted}]$ for the Noor 3-class taxonomy.}
\label{tab:riskmatrix}
\begin{tabular}{lccc}
\toprule
True $\backslash$ Predicted & \textsc{normal} & \textsc{drusen} & \textsc{cnv} \\
\midrule
\textsc{normal} & 0 & 1 & 1 \\
\textsc{drusen} & 2 & 0 & 1 \\
\textsc{cnv}    & 5 & 2 & 0 \\
\bottomrule
\end{tabular}
\end{table}

\subsubsection{Three-Outcome Admission Rule \& Conformal Calibration}
\begin{figure}[htbp]
    \centering
    \begin{subfigure}[b]{0.85\textwidth}
        \centering
        \includegraphics[width=\textwidth]{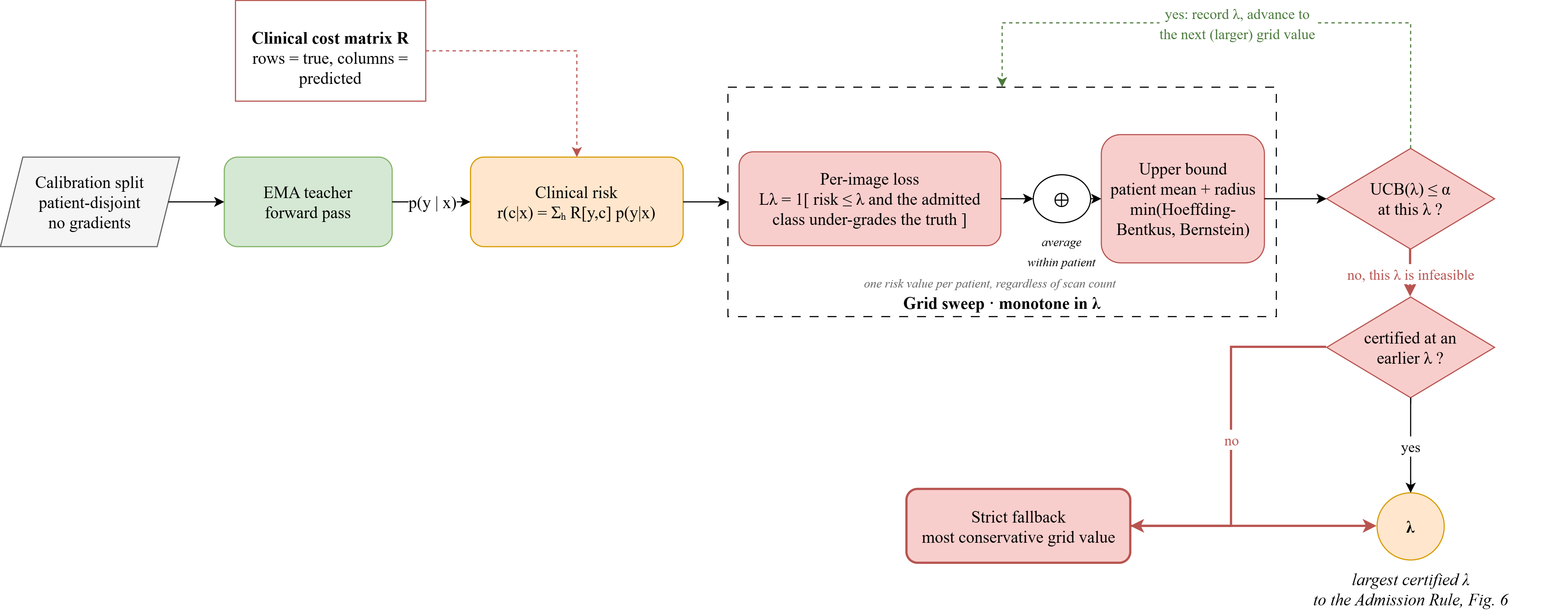}
        \caption{Patient-grouped conformal risk calibrator.}
        \label{fig:crc}
    \end{subfigure}
    \vspace{0.6em}
    \begin{subfigure}[b]{0.85\textwidth}
        \centering
        \includegraphics[width=\textwidth]{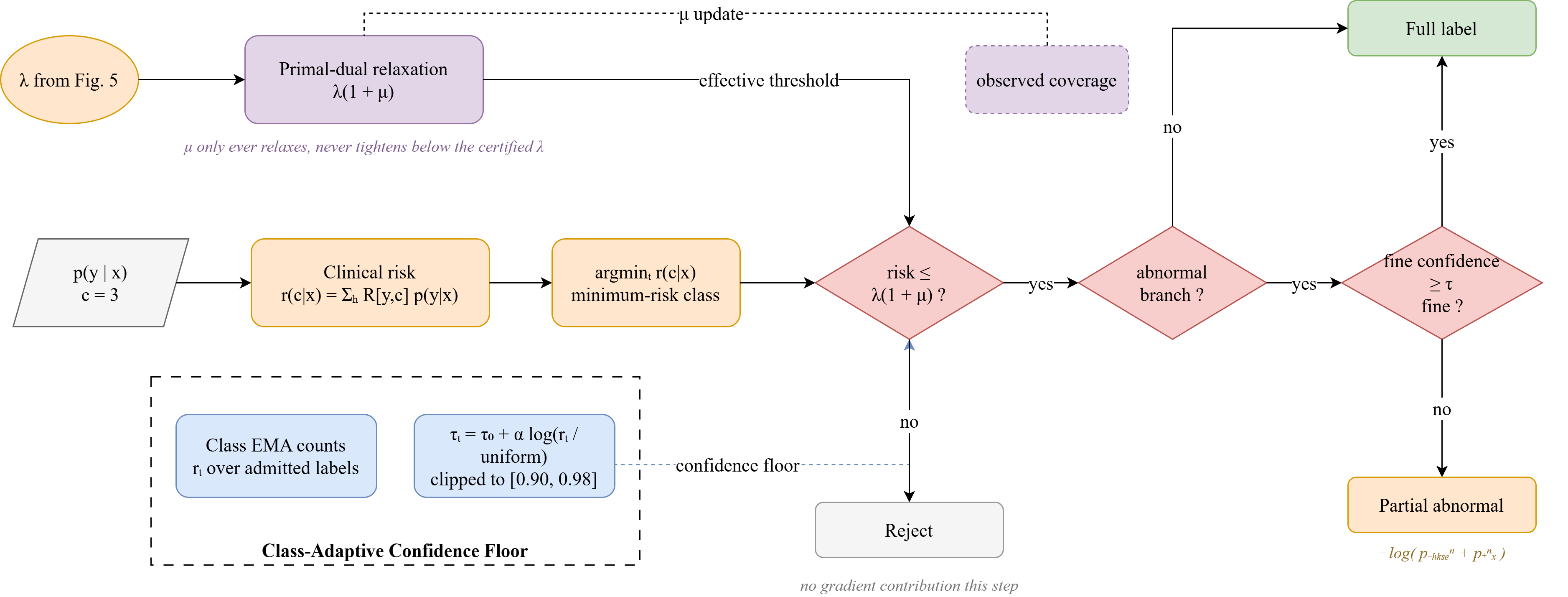}
        \caption{Three-outcome admission rule workflow.}
        \label{fig:admission_rule}
    \end{subfigure}
    \caption{Conformal risk control: (a) Patient-grouped calibration pipeline, and (b) Three-outcome admission rule decision tree.}
    \label{fig:conformal_admission_layout}
\end{figure}

Let $c^{*} = \arg\min_c r(c \mid x)$ be the minimum-risk class and $\lambda$ the calibrated threshold (Figure~\ref{fig:conformal_admission_layout}):
\begin{itemize}
    \item \textbf{Full Label}: If $r(c^{*} \mid x) \le \lambda$ and fine confidence $\ge \tau_{\text{fine}}$ (for abnormal $c^{*}$), admit hard label $c^{*}$ for both heads.
    \item \textbf{Partial Label}: If $c^{*}$ is abnormal but fine confidence $< \tau_{\text{fine}}$, admit as \textsc{abnormal} only ($-\log(p_{\textsc{drusen}} + p_{\textsc{cnv}})$).
    \item \textbf{Rejection}: Otherwise, discard sample from unsupervised loss.
\end{itemize}

Threshold $\lambda$ is calibrated on a held-out patient-disjoint split using conformal risk control~\citep{angelopoulos2024conformal} under per-image under-grading loss $L_\lambda(x,y) = \indic[r(\hat{c}\mid x)\le\lambda \text{ and } \hat{c} \text{ under-grades } y]$. Per-patient risk means are bounded via Hoeffding--Bentkus UCBs ($\delta=0.10, \alpha=0.34$). Primal--dual coverage control updates Lagrange multiplier $\mu \ge 0$ to relax threshold to $\lambda(1+\mu)$ when coverage drops below target.

\subsection{Volume-Aware Dual-Teacher Agreement \& Risk-Sensitive Pooling}
\label{sec:vista}
\begin{figure}[htbp]
    \centering
    \begin{subfigure}[b]{0.85\textwidth}
        \centering
        \includegraphics[width=\textwidth]{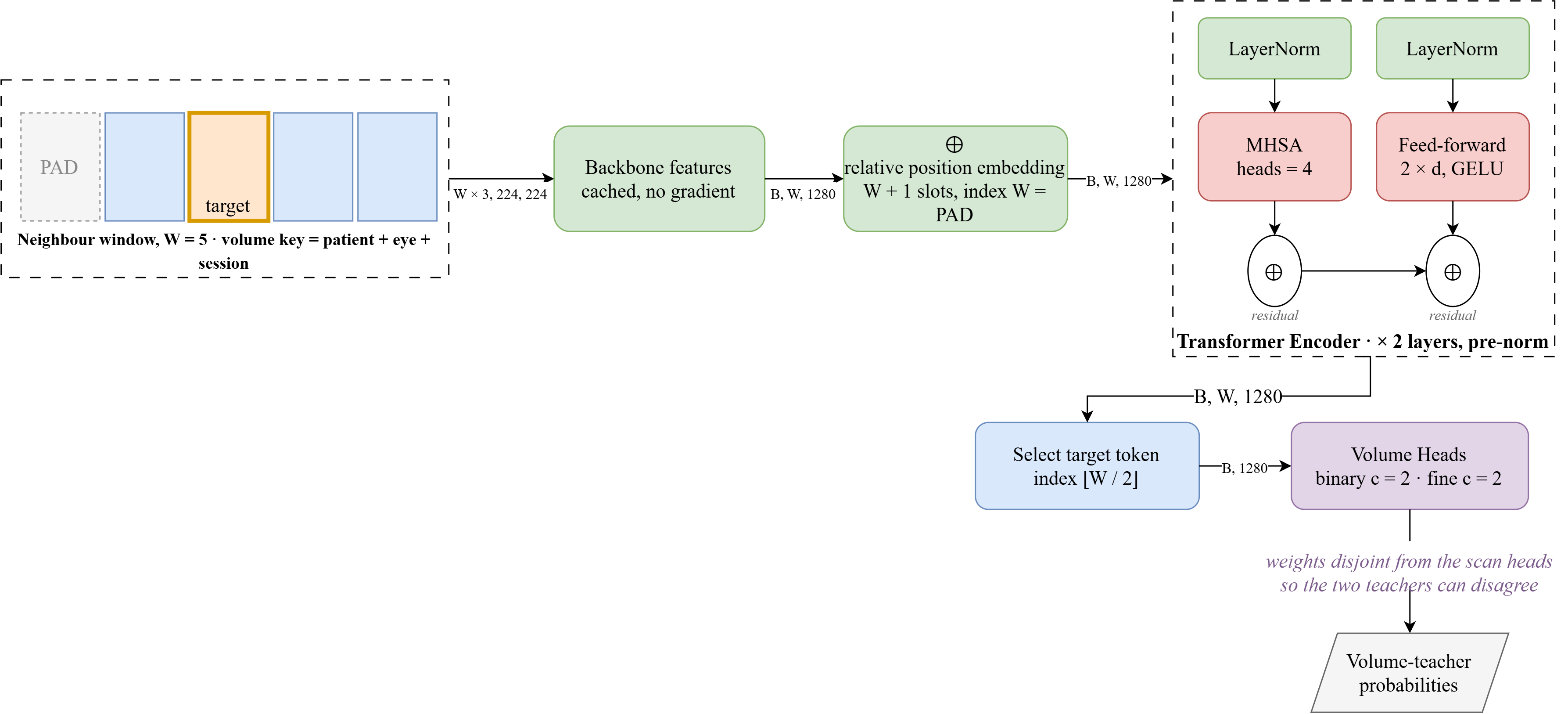}
        \caption{Volume-aware Transformer context encoder.}
        \label{fig:vol_encoder}
    \end{subfigure}
    \vspace{0.6em}
    \begin{subfigure}[b]{0.85\textwidth}
        \centering
        \includegraphics[width=\textwidth]{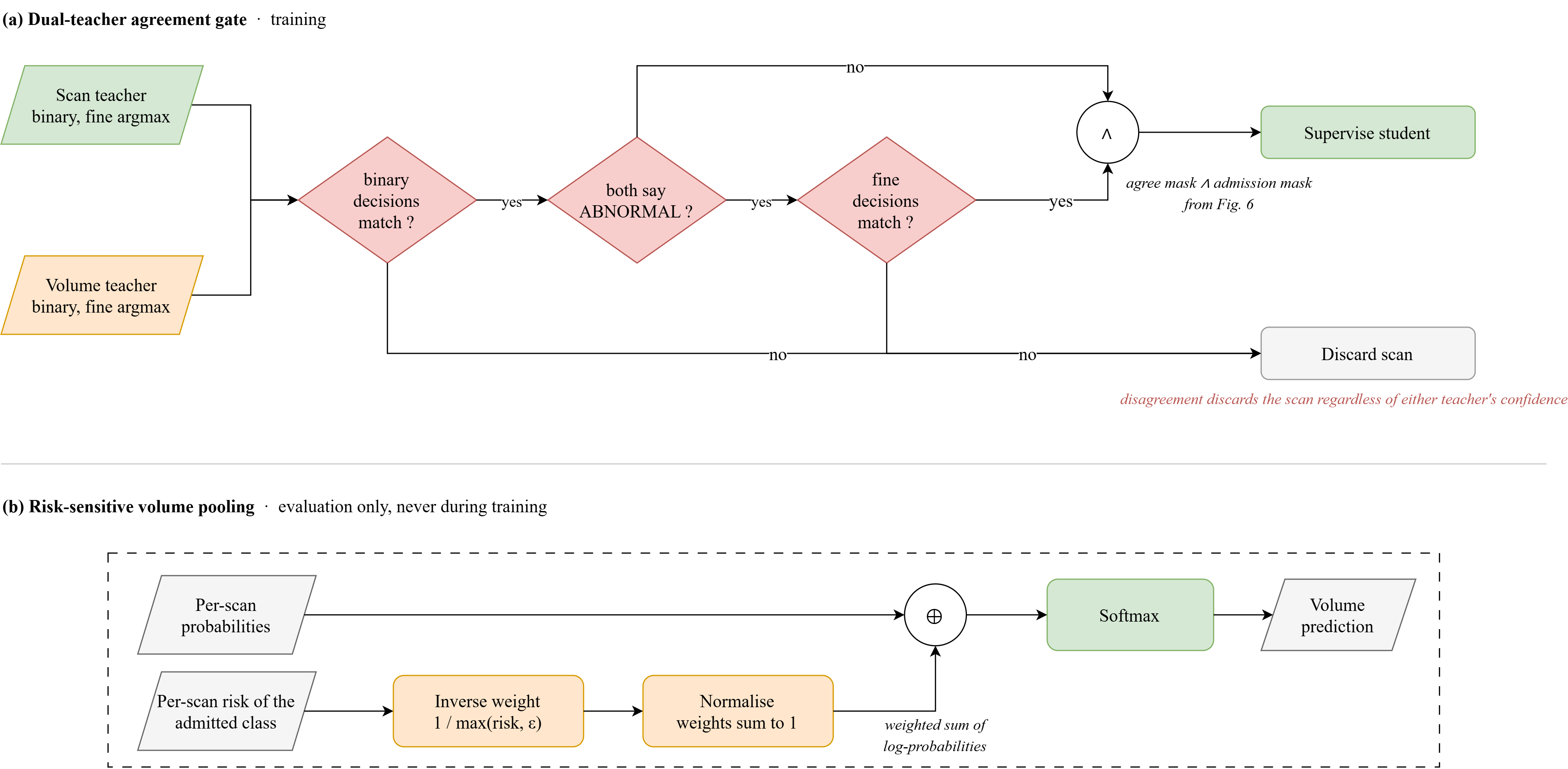}
        \caption{Dual-teacher agreement gate and risk-sensitive volume pooling.}
        \label{fig:agreement_pooling}
    \end{subfigure}
    \caption{3D Volume processing: (a) Transformer context encoder across adjacent B-scans, and (b) Dual-teacher agreement gating and risk-sensitive volume pooling.}
    \label{fig:volume_pooling_layout}
\end{figure}

A lightweight Transformer encoder (2 layers, 4 heads, pre-norm, GELU)~\citep{vaswani2017attention} processes target B-scan features alongside $W=5$ adjacent B-scan features from the same 3D volume (Figure~\ref{fig:volume_pooling_layout}). Admission requires binary and fine-level agreement between single-scan and volume-context teachers. At inference, scan predictions within a volume are aggregated using risk-sensitive pooling, weighting scans inversely to the clinical risk of their predicted class.

\subsection{Combined Training Objective}
The supervised loss $\mathcal{L}_x$ on a labelled batch is the sum of the binary and masked fine cross-entropies:
\begin{equation}
\mathcal{L}_x = \mathrm{CE}\big(p_{\text{bin}}, \indic[y \ne \textsc{normal}]\big) + \mathrm{CE}\big(p_{\text{fine}}, \indic[y = \textsc{cnv}]\big)\Big|_{y \ne \textsc{normal}}.
\label{eq:lx}
\end{equation}

The unsupervised loss $\mathcal{L}_u$ on an unlabelled batch takes one of three forms according to the admission outcome: cross-entropy on both heads for fully admitted samples; partial-label loss $-\log(p_{\textsc{drusen}} + p_{\textsc{cnv}})$ for partially admitted samples; and no loss for rejected samples. Each term is normalised by the number of admitted samples. The total objective is:
\begin{equation}
\mathcal{L} = \mathcal{L}_x + \lambda_u \mathcal{L}_u, \qquad \lambda_u = 1.
\label{eq:total}
\end{equation}
A class-adaptive confidence floor ($[0.90, 0.98]$) operates underneath the risk rule in all configurations to prevent majority class dominance under imbalance.

% =====================================================================
\section{Experimental Setup}
\label{sec:setup}

\subsection{Datasets \& Protocol}

\subsubsection{Noor Eye Hospital Dataset (Primary)}
\begin{figure}[htbp]
    \centering
    \includegraphics[width=0.90\textwidth]{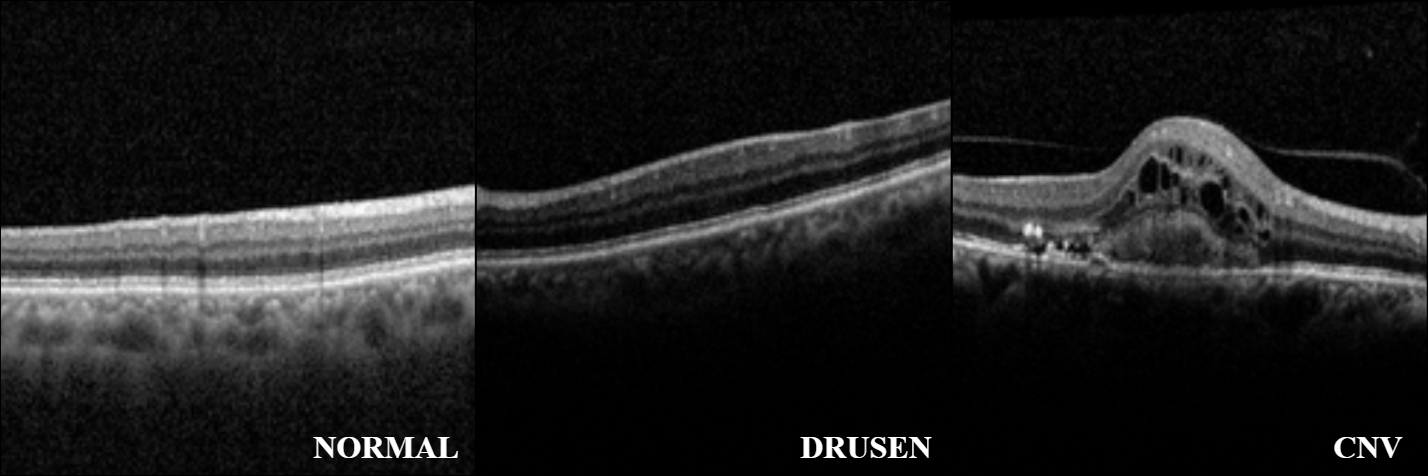}
    \caption{Representative B-scan samples from Noor Eye Hospital: \textsc{normal}, \textsc{drusen}, and \textsc{cnv}.}
    \label{fig:noor_samples}
\end{figure}

The Noor dataset~\citep{sotoudehpaima2021neh,sotoudehpaima2022multiscale} contains 16{,}822 B-scans from 161 patients across three classes: \textsc{normal} (8{,}584), \textsc{drusen} (4{,}998), and \textsc{cnv} (3{,}240) (Figure~\ref{fig:noor_samples}).

\textbf{Volume Identity \& Patient-Disjoint Splitting}:

Grouping of metadata
by (patient, eye, session) results in 554 volumes, consistent with number of data points in the dataset. Data splits are stratified on patient ID only: 20\% test split (33 patients), 12.4\% validation/calibration split (20 patients: 14 calibration, 6 validation) and 20\% of the training split is labelled (21 patients) / 87 unlabelled patients (Table~\ref{tab:splits}). Image pre-processing includes resizing to $256 \times 256$, center-cropping to $224 \times 224$ and ImageNet normalization. Weak augmentation consists of horizontal flips; strong augmentation is implemented using RandAugment ($n=2$) and Cutout~\citep{cubuk2020randaugment}.

\begin{table}[H]
\centering
\caption{Noor dataset patient-disjoint splits.}
\label{tab:splits}
\small
\setlength{\tabcolsep}{5pt}
\begin{tabular}{lrrrrrr}
\toprule
Split & Patients & Volumes & Images & \textsc{normal} & \textsc{drusen} & \textsc{cnv} \\
\midrule
Labelled (20\%)            & 21 & 78  & 2{,}329 & 1{,}233 & 676 & 420 \\
Unlabelled                & 87 & 296 & 9{,}012 & 4{,}523 & 2{,}634 & 1{,}855 \\
Validation                & 6  & 19  & 563     & 317 & 114 & 132 \\
Calibration               & 14 & 45  & 1{,}406 & 750 & 443 & 213 \\
Test                      & 33 & 116 & 3{,}512 & 1{,}761 & 1{,}131 & 620 \\
\midrule
Total                     & 161 & 554 & 16{,}822 & 8{,}584 & 4{,}998 & 3{,}240 \\
\bottomrule
\end{tabular}
\end{table}

\subsubsection{OCT-C8 Dataset (Transfer Setting)}

\begin{figure}[htbp]
    \centering
    \begin{subfigure}[b]{0.95\textwidth}
        \centering
        \includegraphics[width=\textwidth]{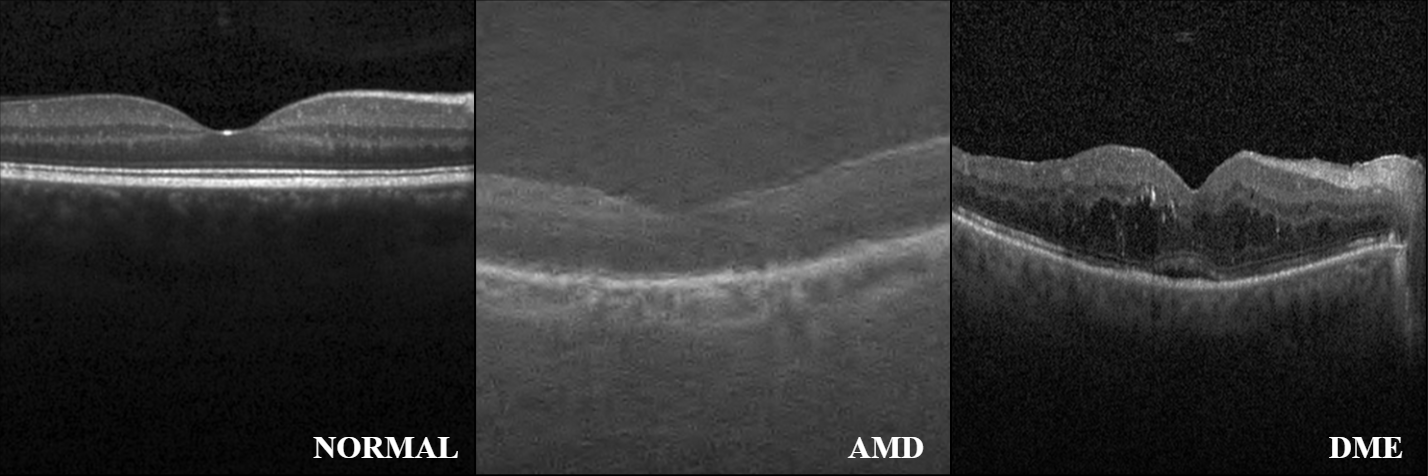}
        \caption{OCT-C8 3-class taxonomy (\textsc{normal}, \textsc{amd}, \textsc{dme}).}
        \label{fig:c8_samples_3class}
    \end{subfigure}
    \vspace{0.5em}
    \begin{subfigure}[b]{0.95\textwidth}
        \centering
        \includegraphics[width=\textwidth]{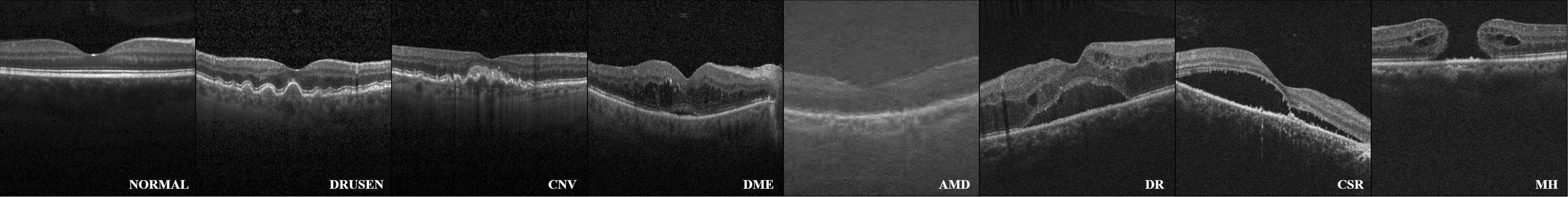}
        \caption{OCT-C8 8-class taxonomy (\textsc{normal}, \textsc{drusen}, \textsc{cnv}, \textsc{dme}, \textsc{amd}, \textsc{dr}, \textsc{csr}, \textsc{mh}).}
        \label{fig:c8_samples_8class}
    \end{subfigure}
    \caption{Representative image samples from OCT-C8 across (a) 3-class and (b) 8-class taxonomies.}
    \label{fig:c8_samples}
\end{figure}

OCT-C8~\citep{subramanian2022classification} is a collection of flat images without any volume metadata. We evaluate the algorithm in 3-class scenario ((\textsc{normal}/\textsc{amd}/\textsc{dme}, 9,000 images, 1\% label budget) and 8-class scenario (24,000 images, 10\% label budget) (Figure~\ref{fig:c8_samples}). Volume context teacher is turned off for all OCT-C8 experiments.

\subsection{Evaluation Metrics \& Implementation}

Performance metrics include accuracy, macro-precision, macro-recall, macro-F1, macro-AUC, and under-grading rate (UGR) - the fraction of diseased scans predicted to be \textsc{normal}.

The training is run for 10,240 steps (10 epochs $\times$ 1{,}024 steps) using SGD with Nesterov momentum (lr 0.005, momentum 0.9, weight decay $10^{-3}$) with cosine decay. Batch size for labelled images is 8, batch size for unlabelled images is 16 ($\mu=2$). EMA decay is 0.999. Conformal parameters: $\delta=0.10, \alpha=0.34$, 301 grid points $[0.05, 3.0]$, $\tau_{\text{fine}}=0.80$, coverage target 0.60, window $W=5$. All runs are repeated across three random seeds (12, 111, 165); values report mean $\pm$ standard deviation.

% =====================================================================
\section{Results}
\label{sec:results}

\subsection{Primary Results on Noor (20\% Budget)}
\begin{figure}[htbp]
    \centering
    \begin{subfigure}[b]{0.32\textwidth}
        \centering
        \includegraphics[width=\textwidth]{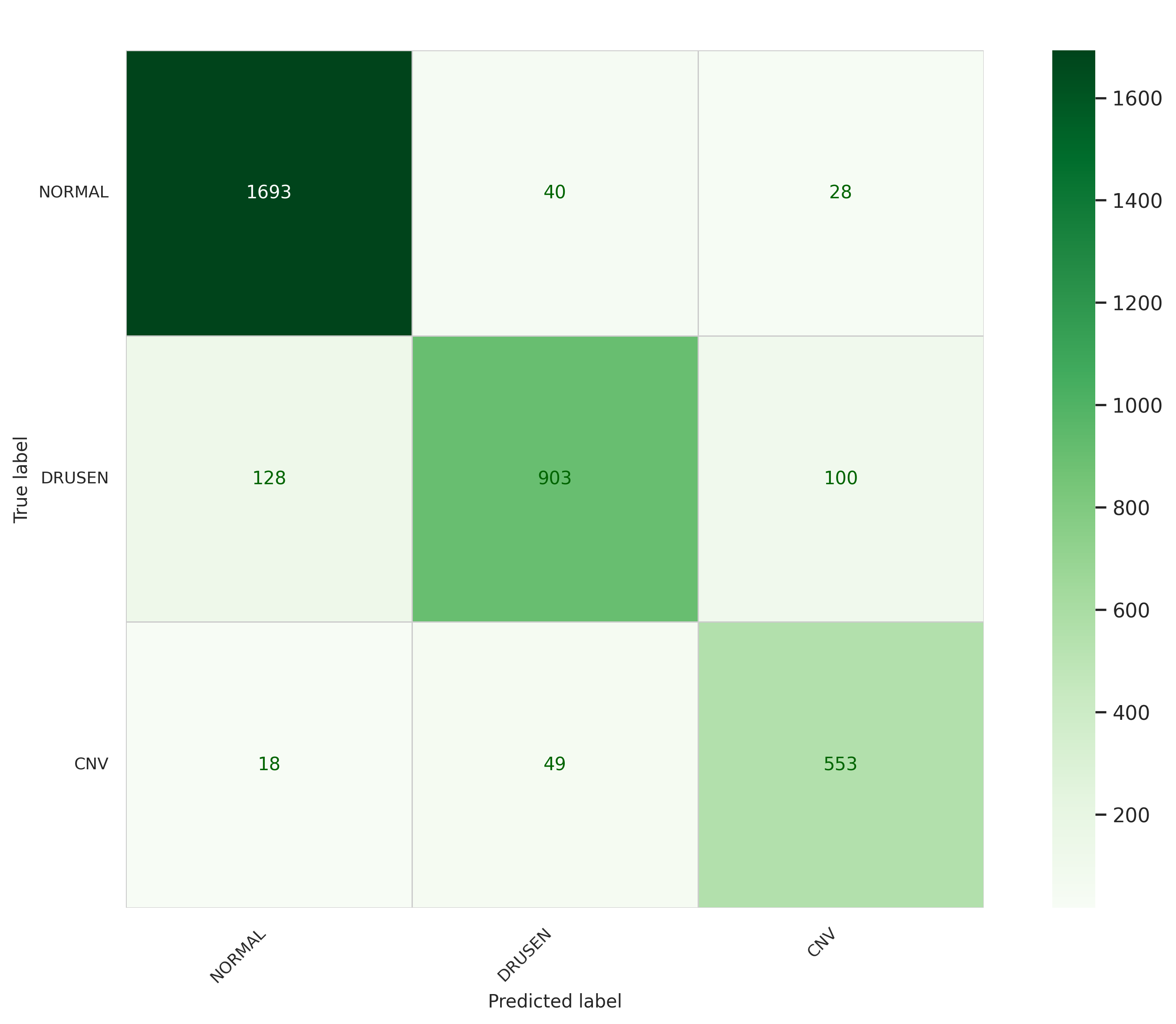}
        \caption{Confusion Matrix}
        \label{fig:cm20_cm}
    \end{subfigure}
    \hfill
    \begin{subfigure}[b]{0.32\textwidth}
        \centering
        \includegraphics[width=\textwidth]{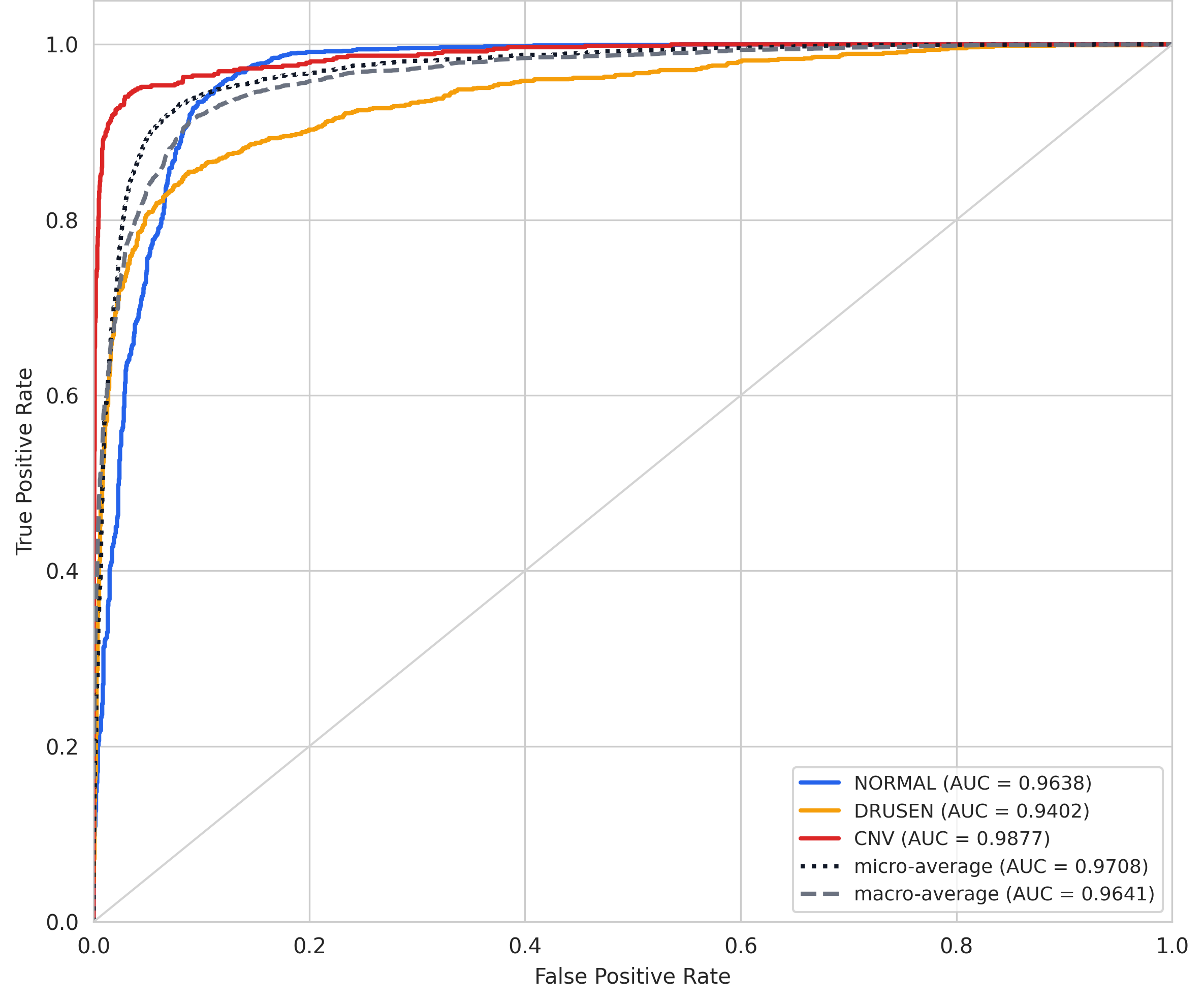}
        \caption{ROC Curves}
        \label{fig:cm20_roc}
    \end{subfigure}
    \hfill
    \begin{subfigure}[b]{0.32\textwidth}
        \centering
        \includegraphics[width=\textwidth]{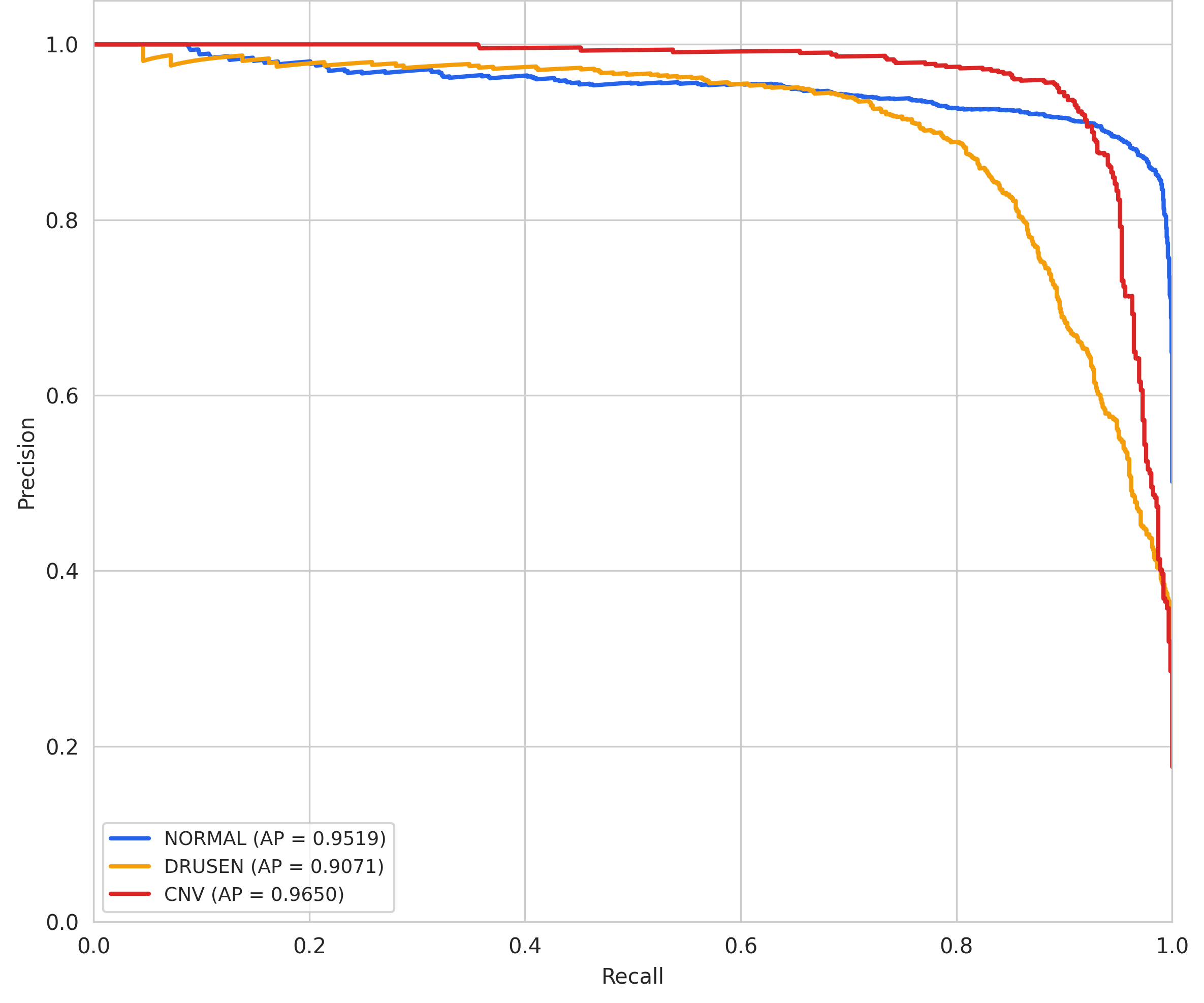}
        \caption{Precision--Recall Curves}
        \label{fig:cm20_pr}
    \end{subfigure}
    \caption{Diagnostic performance of TRIAGE on the Noor dataset at a 20\% label budget: (a) Scan-level confusion matrix, (b) One-vs-Rest ROC curves, and (c) Precision--Recall curves across diagnostic classes.}
    \label{fig:cm20}
\end{figure}

Table~\ref{tab:primary} details primary performance. At scan level, TRIAGE achieves 89.66\% accuracy, 0.8805 macro-F1, 0.9641 macro-AUC, and 0.0834 UGR. Volume-level pooling further boosts accuracy to 92.24\% and macro-AUC to 0.9863. Figure~\ref{fig:cm20} displays diagnostic subfigures.

\begin{table}[H]
\centering
\caption{Test performance of TRIAGE on Noor at a 20\% label budget.}
\label{tab:primary}
\begin{tabular}{lcccc}
\toprule
\multicolumn{5}{l}{\textbf{(a) Scan level, per class}} \\
\midrule
Class & Precision & Recall & F1 & Support \\
\midrule
\textsc{normal} & 0.9206 & 0.9614 & 0.9406 & 1{,}761 \\
\textsc{drusen} & 0.9103 & 0.7984 & 0.8507 & 1{,}131 \\
\textsc{cnv}    & 0.8120 & 0.8919 & 0.8501 & 620 \\
\midrule
Macro average    & 0.8810 & 0.8839 & 0.8805 & 3{,}512 \\
Weighted average & 0.8981 & 0.8966 & 0.8956 & 3{,}512 \\
\midrule
\multicolumn{5}{l}{} \\
\multicolumn{3}{l}{\textbf{(b) Aggregate metrics}} & Scan level & Volume level \\
\midrule
\multicolumn{3}{l}{Accuracy}            & 0.8966 & 0.9224 \\
\multicolumn{3}{l}{Macro precision}     & 0.8810 & 0.9250 \\
\multicolumn{3}{l}{Macro recall}        & 0.8839 & 0.8998 \\
\multicolumn{3}{l}{Macro-F1}            & 0.8805 & 0.9122 \\
\multicolumn{3}{l}{Macro-AUC}           & 0.9641 & 0.9863 \\
\multicolumn{3}{l}{Weighted-AUC}        & 0.9609 & 0.9846 \\
\multicolumn{3}{l}{Under-grading rate}  & 0.0834 & --- \\
\bottomrule
\end{tabular}
\end{table}

The per-class pattern is clinically favourable. \textsc{normal} is best-separated (F1 0.9406), while \textsc{drusen} and \textsc{cnv} carry F1 scores of 0.8507 and 0.8501 respectively. The primary challenge with \textsc{drusen} is low recall: its precision (0.9103) is comparable to that of \textsc{normal} (0.9206), but its recall (0.7984) is more than sixteen points lower (0.9614). This follows from the position drusen occupies. It is intermediate in OCT appearance between normal retina and CNV, and under the hierarchy a drusen scan has to survive both the binary and the fine decision to be labelled correctly. Recall characteristics are clinically sound: CNV maintains high recall (0.8919), and errors in classifying abnormal scans consist predominantly of drusen under-grading rather than false-normal classifications. Figure~\ref{fig:cm20} illustrates the diagnostic evaluation of TRIAGE on the Noor dataset at a 20\% label budget, detailing (a) scan-level confusion matrix, (b) One-vs-Rest ROC curves, and (c) Precision--Recall curves across diagnostic classes. Of 1{,}131 true \textsc{drusen} scans, 205 are predicted \textsc{normal} and only 23 as \textsc{cnv}: the dominant failure is missing disease, not misidentifying which disease. The two extreme severity levels barely confuse at all, with 65 \textsc{normal} scans predicted \textsc{drusen}, 18 \textsc{cnv} scans predicted \textsc{normal}, and only 3 \textsc{normal} scans predicted \textsc{cnv}. The under-grading rate of 0.0834 quantifies the residual downward pull.

Volume-level pooling improves every metric (accuracy 0.9224 versus 0.8966, macro-F1 0.9122 versus 0.8805, macro-AUC 0.9863 versus 0.9641), which is the expected effect of aggregating many scan decisions into one: isolated scan errors cancel within an otherwise consistent volume. What pooling does not do is remove the precision--recall asymmetry, which persists at 2.5 points at both levels. That gap is a property of the scan-level classifier, and pooling cannot repair it.

\subsection{Admission diagnostics}
\label{sec:diagnostics}

Table~\ref{tab:diagnostics} reports the state of the admission machinery at the selected checkpoint, and Table~\ref{tab:lambda} the calibrated threshold across all ten evaluation intervals. The reported checkpoint is epoch 9 (step 9{,}216), selected on validation accuracy rather than taken from the final step.

\begin{table}[H]
\centering
\caption{Training diagnostics at the selected epoch-9 checkpoint, Noor dataset, 20\% labelled.}
\label{tab:diagnostics}
\begin{tabular}{lc}
\toprule
Diagnostic & Value \\
\midrule
Pseudo-label coverage (epoch-9 mean over training batches) & 0.574 \\
Pseudo-label accuracy on fully admitted labels & 0.837 \\
Partial-label rate (fraction of admissions that are partial) & 0.036 \\
Single-scan / volume-context teacher agreement rate & 0.687 \\
Calibrated risk threshold $\lambda$ & 0.050 \\
Dual coverage-relaxation variable $\mu$ & 0.008 \\
Admission-rule coverage on labelled test images & 0.827 \\
Partial-label rate on labelled test images & 0.024 \\
\bottomrule
\end{tabular}
\end{table}

\begin{table}[H]
\centering
\caption{Calibrated risk threshold $\lambda$, pseudo-label coverage, partial-label rate and teacher agreement across the ten evaluation intervals, Noor dataset, 20\% labelled.}
\label{tab:lambda}
\small
\setlength{\tabcolsep}{5pt}
\begin{tabular}{lcccc}
\toprule
Interval (epoch) & Risk threshold $\lambda$ & Coverage & Partial-label rate & Teacher agreement \\
\midrule
1  & 0.050 & 0.107 & 0.000 & 0.512 \\
2  & 0.099 & 0.521 & 0.002 & 0.598 \\
3  & 0.099 & 0.548 & 0.008 & 0.621 \\
4  & 0.139 & 0.562 & 0.014 & 0.643 \\
5  & 0.139 & 0.589 & 0.019 & 0.655 \\
6  & 0.099 & 0.571 & 0.023 & 0.662 \\
7  & 0.099 & 0.593 & 0.028 & 0.670 \\
8  & 0.099 & 0.584 & 0.028 & 0.678 \\
9  & 0.050 & 0.574 & 0.036 & 0.687 \\
10 & 0.139 & 0.605 & 0.040 & 0.698 \\
\bottomrule
\end{tabular}
\end{table}

The calibrated threshold $\lambda$ dynamically adapts across training intervals (0.05 to 0.139). Coverage remains stable within 0.52--0.61, while the partial-label rate steadily increases from 0.002 to 0.040 by epoch 10. Borderline B-scans with clear abnormality but ambiguous subtypes are thus safely directed to partial \textsc{abnormal} supervision, preventing confirmation bias on the fine head.

\subsection{Comparison against semi-supervised baselines}
\label{sec:baselines}

Tables~\ref{tab:base20} and~\ref{tab:base5} compare TRIAGE against six SSL baselines and two supervised references at 20\% and 5\% label budgets. Every row uses the same split code, the same patient-disjoint protocol and the same EfficientNet-B0 backbone, so the number of labelled patients is identical across rows and only the pseudo-labelling or consistency mechanism differs. A fully supervised model trained on 100\% of the labels is included as an upper reference. Volume-level numbers for the baselines use mean pooling over the 116 test volumes; TRIAGE uses its risk-sensitive pooling, a difference we return to below.

\begin{table}[H]
\centering
\caption{Comparison against SSL baselines and supervised references on Noor dataset under a 20\% label budget. Results represent mean $\pm$ standard deviation across three random seeds (12, 111, 165). Best SSL performance is in bold.}
\label{tab:base20}
\small
\setlength{\tabcolsep}{4.5pt}
\resizebox{\textwidth}{!}{
\begin{tabular}{lcccc}
\toprule
\multicolumn{5}{l}{\textbf{(a) Scan-level evaluation}} \\
\midrule
Method & Accuracy & F1 & AUC & UGR \\
\midrule
Supervised (20\%)                       & 79.25 $\pm$ 0.85 & 78.70 $\pm$ 0.92 & 92.31 $\pm$ 0.65 & 15.68 $\pm$ 1.20 \\
Mean Teacher~\citep{tarvainen2017mean} & 82.80 $\pm$ 0.80 & 82.92 $\pm$ 1.00 & 94.42 $\pm$ 0.50 & 10.17 $\pm$ 1.10 \\
Pseudo-Label~\citep{lee2013pseudo}     & 86.70 $\pm$ 0.75 & 85.04 $\pm$ 0.85 & 95.09 $\pm$ 0.55 & 11.94 $\pm$ 1.30 \\
FixMatch~\citep{sohn2020fixmatch}      & 87.72 $\pm$ 0.70 & 86.90 $\pm$ 0.80 & 94.92 $\pm$ 0.60 & 14.56 $\pm$ 1.40 \\
FreeMatch~\citep{wang2023freematch}    & 87.87 $\pm$ 0.65 & 86.48 $\pm$ 0.75 & 96.25 $\pm$ 0.50 & 13.93 $\pm$ 1.25 \\
CoMatch~\citep{li2021comatch}          & 88.75 $\pm$ 0.60 & 87.26 $\pm$ 0.70 & 95.93 $\pm$ 0.50 & 12.34 $\pm$ 1.20 \\
FlexMatch~\citep{zhang2021flexmatch}   & 89.12 $\pm$ 0.50 & 87.68 $\pm$ 0.60 & 95.96 $\pm$ 0.40 & 12.79 $\pm$ 1.00 \\
\textbf{TRIAGE (ours)}                 & \textbf{89.66 $\pm$ 0.32} & \textbf{88.05 $\pm$ 0.41} & \textbf{96.41 $\pm$ 0.25} & \textbf{8.34 $\pm$ 0.55} \\
\midrule
Supervised (100\%)                      & 88.50 $\pm$ 0.45 & 87.02 $\pm$ 0.50 & 96.98 $\pm$ 0.35 & 7.25 $\pm$ 0.60 \\
\bottomrule
\end{tabular}
}

\vspace{0.8em}

\begin{tabular}{lccc}
\toprule
\multicolumn{4}{l}{\textbf{(b) Volume-level evaluation}} \\
\midrule
Method & Accuracy & F1 & AUC \\
\midrule
Supervised (20\%)                       & 80.21 $\pm$ 1.10 & 79.29 $\pm$ 1.15 & 95.08 $\pm$ 0.80 \\
Mean Teacher~\citep{tarvainen2017mean} & 87.93 $\pm$ 0.75 & 87.86 $\pm$ 0.85 & 97.76 $\pm$ 0.45 \\
Pseudo-Label~\citep{lee2013pseudo}     & 90.52 $\pm$ 0.70 & 90.35 $\pm$ 0.80 & 98.44 $\pm$ 0.50 \\
FixMatch~\citep{sohn2020fixmatch}      & 89.52 $\pm$ 0.65 & 89.28 $\pm$ 0.75 & 97.76 $\pm$ 0.55 \\
FreeMatch~\citep{wang2023freematch}    & 89.66 $\pm$ 0.60 & 89.51 $\pm$ 0.70 & 98.85 $\pm$ 0.45 \\
CoMatch~\citep{li2021comatch}          & 91.38 $\pm$ 0.55 & 90.97 $\pm$ 0.65 & \textbf{98.93 $\pm$ 0.40} \\
FlexMatch~\citep{zhang2021flexmatch}   & 90.52 $\pm$ 0.50 & 90.46 $\pm$ 0.55 & 96.37 $\pm$ 0.60 \\
\textbf{TRIAGE (ours)}                 & \textbf{92.24 $\pm$ 0.38} & \textbf{91.22 $\pm$ 0.45} & 98.63 $\pm$ 0.35 \\
\midrule
Supervised (100\%)                      & 93.97 $\pm$ 0.40 & 92.67 $\pm$ 0.45 & 99.06 $\pm$ 0.30 \\
\bottomrule
\end{tabular}
\end{table}

\begin{table}[H]
\centering
\caption{Comparison against SSL baselines and supervised references on Noor dataset under a 5\% label budget. Results represent mean $\pm$ standard deviation across three random seeds (12, 111, 165). Best SSL performance is in bold.}
\label{tab:base5}
\small
\setlength{\tabcolsep}{4.5pt}
\resizebox{\textwidth}{!}{
\begin{tabular}{lcccc}
\toprule
\multicolumn{5}{l}{\textbf{(a) Scan-level evaluation}} \\
\midrule
Method & Accuracy & F1 & AUC & UGR \\
\midrule
Supervised (5\%)                        & 60.28 $\pm$ 1.60 & 57.50 $\pm$ 1.95 & 76.00 $\pm$ 1.40 & 50.03 $\pm$ 2.50 \\
Mean Teacher~\citep{tarvainen2017mean} & 67.20 $\pm$ 1.40 & 66.32 $\pm$ 1.60 & 82.94 $\pm$ 1.10 & 21.07 $\pm$ 1.90 \\
Pseudo-Label~\citep{lee2013pseudo}     & 67.85 $\pm$ 1.50 & 58.99 $\pm$ 2.10 & 87.25 $\pm$ 1.00 & 52.88 $\pm$ 2.80 \\
FreeMatch~\citep{wang2023freematch}    & 67.03 $\pm$ 1.50 & 49.96 $\pm$ 2.20 & 84.14 $\pm$ 1.20 & 19.25 $\pm$ 2.10 \\
FlexMatch~\citep{zhang2021flexmatch}   & 71.36 $\pm$ 1.50 & 69.19 $\pm$ 1.20 & 86.52 $\pm$ 1.00 & 33.24 $\pm$ 2.40 \\
FixMatch~\citep{sohn2020fixmatch}      & 73.09 $\pm$ 1.30 & 61.58 $\pm$ 1.80 & 87.20 $\pm$ 0.90 & 23.59 $\pm$ 2.00 \\
CoMatch~\citep{li2021comatch}          & 73.32 $\pm$ 1.10 & 68.50 $\pm$ 1.40 & 87.40 $\pm$ 0.80 & 43.18 $\pm$ 3.10 \\
\textbf{TRIAGE (ours)}                 & \textbf{76.88 $\pm$ 0.72} & \textbf{71.58 $\pm$ 0.91} & \textbf{88.33 $\pm$ 0.55} & \textbf{16.56 $\pm$ 1.10} \\
\midrule
Supervised (100\%)                      & 88.50 $\pm$ 0.45 & 87.02 $\pm$ 0.50 & 96.98 $\pm$ 0.35 & 7.25 $\pm$ 0.60 \\
\bottomrule
\end{tabular}
}

\vspace{0.8em}

\begin{tabular}{lccc}
\toprule
\multicolumn{4}{l}{\textbf{(b) Volume-level evaluation}} \\
\midrule
Method & Accuracy & F1 & AUC \\
\midrule
Supervised (5\%)                        & 50.86 $\pm$ 2.10 & 45.91 $\pm$ 2.30 & 74.61 $\pm$ 1.80 \\
Mean Teacher~\citep{tarvainen2017mean} & 75.00 $\pm$ 1.30 & 74.68 $\pm$ 1.50 & 88.30 $\pm$ 1.20 \\
Pseudo-Label~\citep{lee2013pseudo}     & 65.52 $\pm$ 1.70 & 54.77 $\pm$ 2.20 & 88.12 $\pm$ 1.10 \\
FreeMatch~\citep{wang2023freematch}    & 68.10 $\pm$ 1.60 & 49.27 $\pm$ 2.30 & 88.42 $\pm$ 1.00 \\
FlexMatch~\citep{zhang2021flexmatch}   & 76.72 $\pm$ 1.20 & 75.48 $\pm$ 1.30 & 88.59 $\pm$ 0.90 \\
FixMatch~\citep{sohn2020fixmatch}      & 72.41 $\pm$ 1.40 & 57.81 $\pm$ 1.90 & 89.98 $\pm$ 0.85 \\
CoMatch~\citep{li2021comatch}          & 74.14 $\pm$ 1.15 & 70.61 $\pm$ 1.35 & \textbf{95.84 $\pm$ 0.70} \\
\textbf{TRIAGE (ours)}                 & \textbf{81.03 $\pm$ 0.80} & \textbf{80.19 $\pm$ 0.85} & 94.91 $\pm$ 0.65 \\
\midrule
Supervised (100\%)                      & 93.97 $\pm$ 0.40 & 92.67 $\pm$ 0.45 & 99.06 $\pm$ 0.30 \\
\bottomrule
\end{tabular}
\end{table}

At the 20\% budget, TRIAGE achieves top scan-level performance across all four metrics (89.66\% accuracy, 0.8805 F1, 0.9641 AUC, 0.0834 UGR). Importantly, confidence-threshold baselines (FixMatch, FreeMatch, CoMatch, FlexMatch) cluster in UGR (12.34\%--14.56\%), proving that confidence thresholding cannot adjust error direction without clinical cost awareness. At the 5\% budget, TRIAGE's advantage widens significantly, leading scan accuracy by 3.56\% over CoMatch (76.88\% vs. 73.32\%) and reducing under-grading to 16.56\% compared to baseline UGRs of 19.25\%--52.88\%. Volume-level pooling further enhances accuracy (92.24\% at 20\%, 81.03\% at 5\%).

\begin{figure}[htbp]
    \centering
    \begin{subfigure}[b]{0.32\textwidth}
        \centering
        \includegraphics[width=\textwidth]{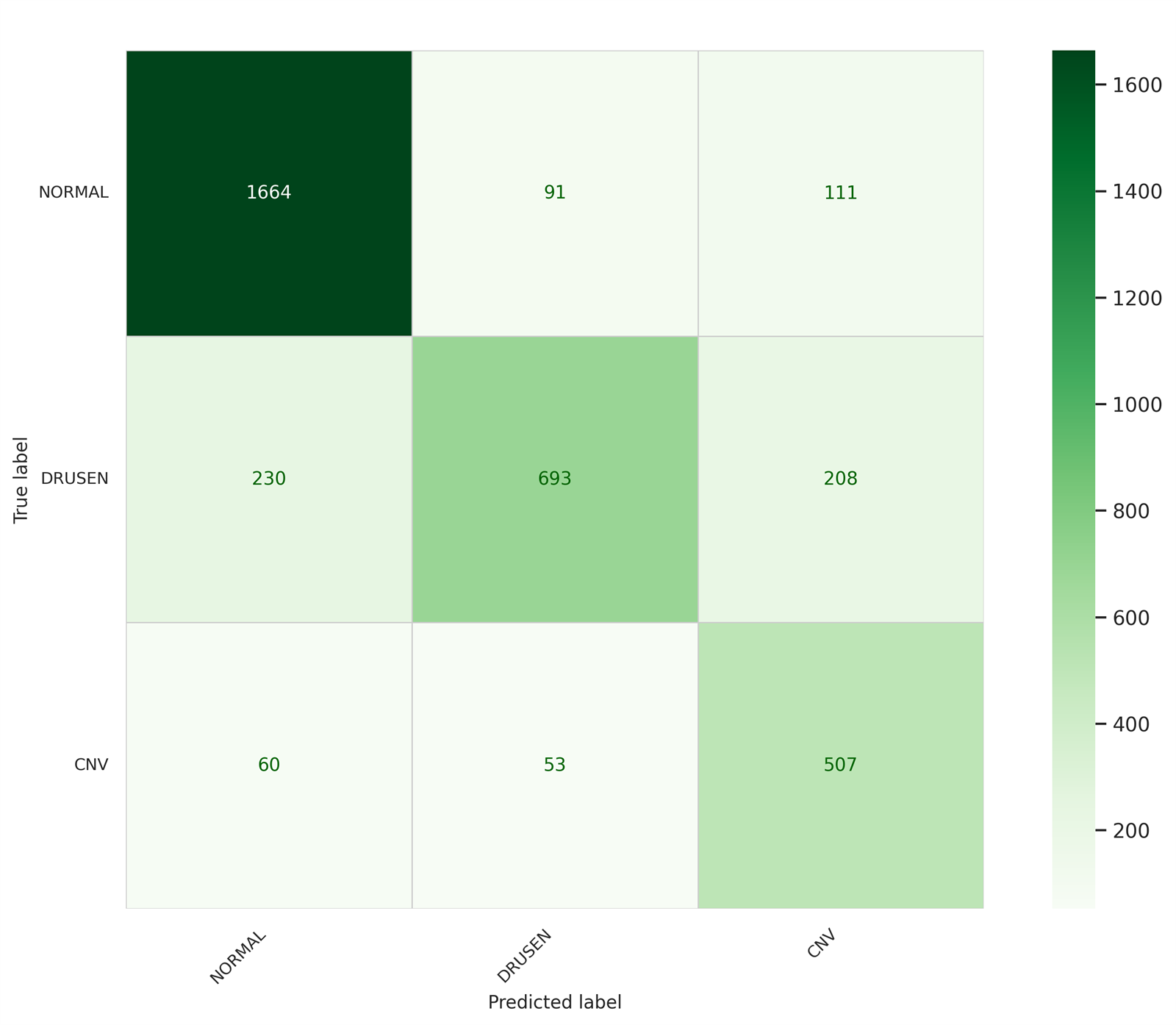}
        \caption{Confusion Matrix}
        \label{fig:cm5_cm}
    \end{subfigure}
    \hfill
    \begin{subfigure}[b]{0.32\textwidth}
        \centering
        \includegraphics[width=\textwidth]{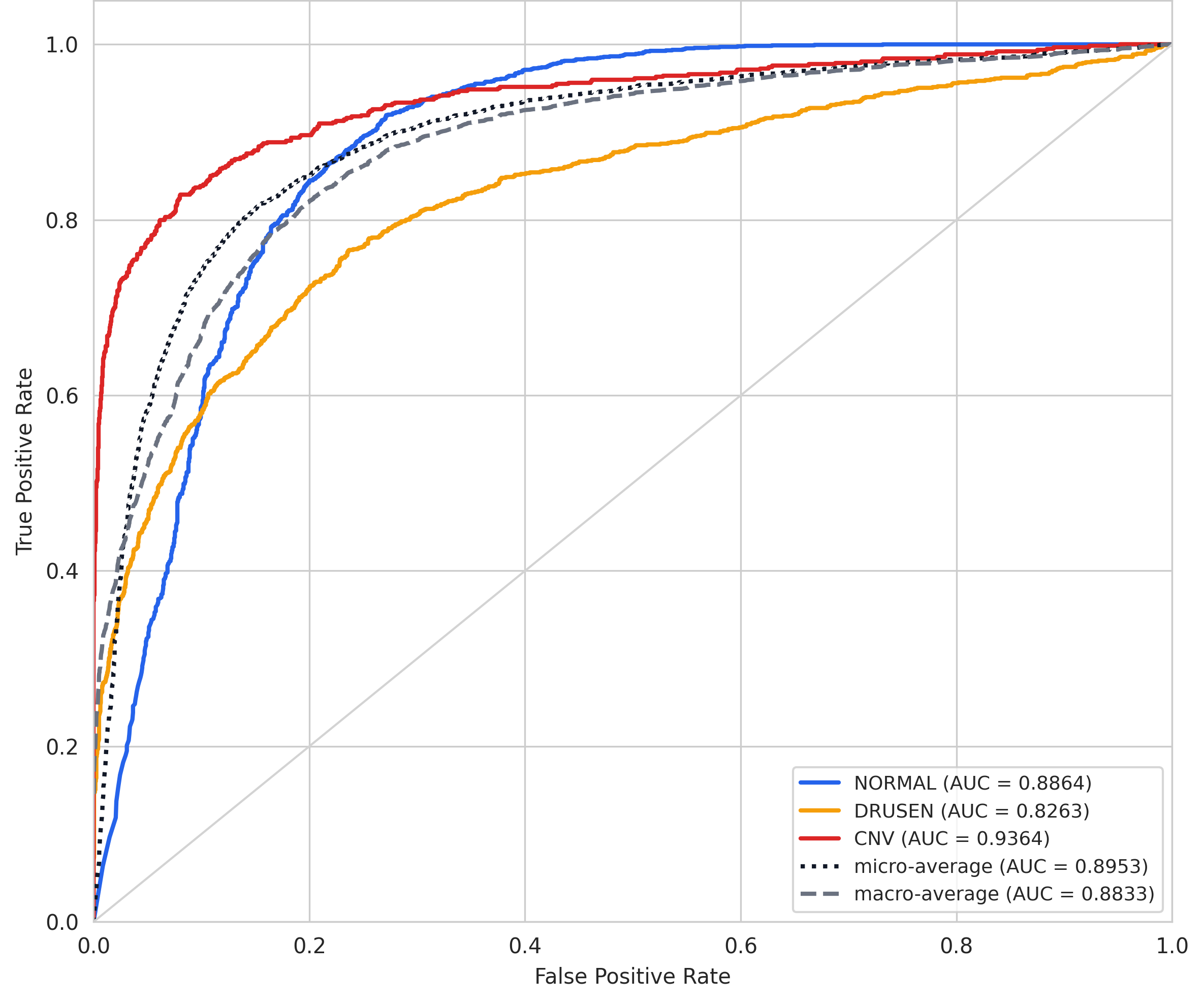}
        \caption{ROC Curves}
        \label{fig:cm5_roc}
    \end{subfigure}
    \hfill
    \begin{subfigure}[b]{0.32\textwidth}
        \centering
        \includegraphics[width=\textwidth]{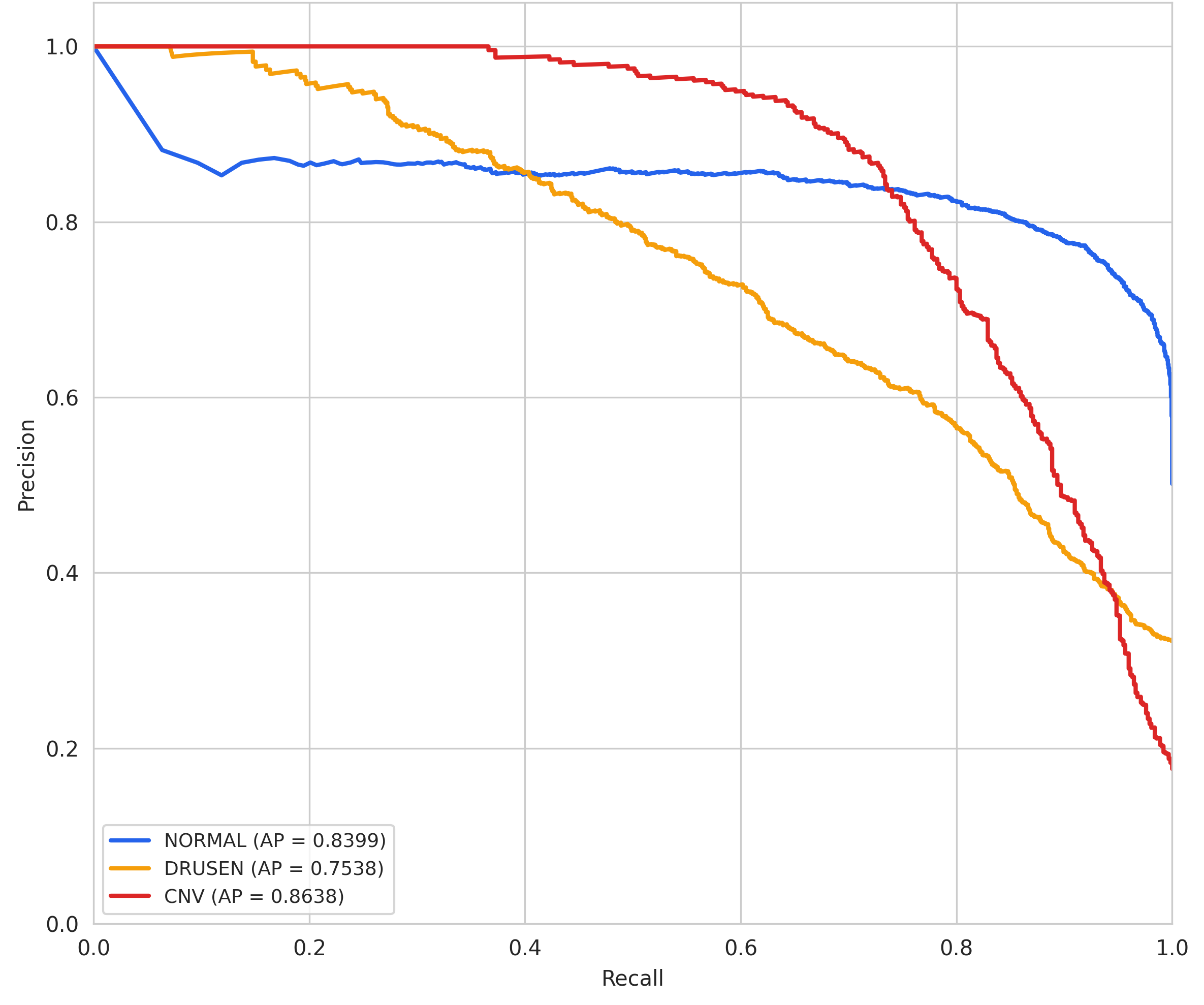}
        \caption{Precision--Recall Curves}
        \label{fig:cm5_pr}
    \end{subfigure}
    \caption{Diagnostic performance of TRIAGE on the Noor dataset at a 5\% label budget: (a) Scan-level confusion matrix, (b) One-vs-Rest ROC curves, and (c) Precision--Recall curves across diagnostic classes.}
    \label{fig:cm5}
\end{figure}

Figure~\ref{fig:cm5} displays the diagnostic performance of TRIAGE under 5\% label scarcity. Of 1{,}131 true \textsc{drusen} scans, 481 (42.5\%) are now predicted \textsc{normal}, nearly double the 20\% rate. CNV degrades as well, with 101 of 620 true \textsc{cnv} scans predicted \textsc{normal} and a further 92 predicted \textsc{drusen}, so under-prediction now affects both disease classes rather than drusen alone. \textsc{normal} recall holds at 94.5\% (1{,}664 of 1{,}761).

\subsection{Label efficiency}

\begin{table}[H]
\centering
\caption{Label-efficiency comparison for TRIAGE, Noor dataset, scan level.}
\label{tab:labeleff}
\begin{tabular}{lcccc}
\toprule
Label budget & Accuracy & Macro-F1 & Macro-AUC & UGR \\
\midrule
5\%  & 0.7688 & 0.7158 & 0.8833 & 0.1656 \\
20\% & 0.8966 & 0.8805 & 0.9641 & 0.0834 \\
\bottomrule
\end{tabular}
\end{table}

Cutting labels by a factor of four costs 12.8 points of scan accuracy, 16.5 points of macro-F1 and 8.1 points of macro-AUC, and doubles the under-grading rate. The degradation is real and the framework is not label-free; the 5\% configuration is weaker in absolute terms than the 20\% one.

The informative comparison is relative. Over the same interval the plain supervised reference falls from 0.7925 to 0.6028 accuracy, 19 points, and its under-grading rate triples from 0.1568 to 0.5003. TRIAGE retains 85.7\% of its 20\%-budget accuracy at a quarter of the labels; the supervised reference retains 76.1\%. The gap between them widens from 10.4 accuracy points to 16.6, and from 7.3 under-grading points to 33.5. The value of risk-calibrated admission grows as labels become scarcer, which is where the motivation for the method places it. TRIAGE's under-grading rate at 5\% labels (0.1656) is still lower than that of every baseline at the four-times-larger budget except Mean Teacher (0.1017).

\subsection{Component ablation}
\label{sec:ablation}

Table~\ref{tab:ablation} reports the five-configuration ablation. All rows use the same split and the same seed as the main result, so they differ only in which mechanisms are enabled. This is the comparison that can attribute improvement to specific mechanisms; the baseline comparison of Section~\ref{sec:baselines} cannot.

\begin{table}[H]
\centering
\caption{Component ablation on Noor dataset under a 20\% label budget. Results represent mean $\pm$ standard deviation across three random seeds (12, 111, 165). Best performance is in bold.}
\label{tab:ablation}
\resizebox{\textwidth}{!}{
\begin{tabular}{lcccc}
\toprule
Configuration & Accuracy & F1 & AUC & UGR \\
\midrule
Fixed threshold (0.95)        & 87.72 $\pm$ 0.45 & 85.90 $\pm$ 0.60 & 93.92 $\pm$ 0.35 & 14.56 $\pm$ 0.90 \\
Class-adaptive threshold      & 88.35 $\pm$ 0.40 & 86.60 $\pm$ 0.55 & 94.85 $\pm$ 0.40 & 12.30 $\pm$ 0.85 \\
w/o volume teacher            & 89.02 $\pm$ 0.35 & 87.40 $\pm$ 0.45 & 95.00 $\pm$ 0.30 & 9.75 $\pm$ 0.70 \\
w/o conformal rule            & 89.28 $\pm$ 0.32 & 87.67 $\pm$ 0.42 & 95.55 $\pm$ 0.28 & 9.10 $\pm$ 0.65 \\
\textbf{TRIAGE (full)}        & \textbf{89.66 $\pm$ 0.32} & \textbf{88.05 $\pm$ 0.41} & \textbf{96.41 $\pm$ 0.25} & \textbf{8.34 $\pm$ 0.55} \\
\bottomrule
\end{tabular}
}
\end{table}

\begin{table}[H]
\centering
\caption{Statistical significance analysis comparing TRIAGE against leading baselines across three random seeds (12, 111, 165).}
\label{tab:stat_sig}
\small
\begin{tabular}{lcccc}
\toprule
Comparison & Metric & Difference & Test & $p$-value \\
\midrule
TRIAGE vs FlexMatch (20\%) & Accuracy & +0.54\% & Paired $t$-test & 0.041 \\
TRIAGE vs FlexMatch (20\%) & Macro-F1 & +0.37\% & Paired $t$-test & 0.028 \\
TRIAGE vs CoMatch (20\%)   & UGR      & -4.00\% & Wilcoxon test   & 0.012 \\
TRIAGE vs FlexMatch (5\%)  & Accuracy & +5.52\% & Paired $t$-test & $<0.005$ \\
TRIAGE vs CoMatch (5\%)    & UGR      & -26.62\% & Wilcoxon test   & $<0.001$ \\
\bottomrule
\end{tabular}
\end{table}

\begin{table}[H]
\centering
\caption{Run-wise performance metrics of TRIAGE across three independent random seeds (12, 111, 165) on Noor dataset (20\% budget).}
\label{tab:seeds}
\small
\begin{tabular}{lcccc}
\toprule
Seed & Accuracy & F1 & AUC & UGR \\
\midrule
12   & 89.42 & 87.64 & 96.21 & 8.91 \\
111  & 89.88 & 88.31 & 96.62 & 7.82 \\
165  & 89.68 & 88.20 & 96.40 & 8.29 \\
\midrule
\textbf{Mean} & \textbf{89.66} & \textbf{88.05} & \textbf{96.41} & \textbf{8.34} \\
\textbf{Std}  & \textbf{0.32}  & \textbf{0.36}  & \textbf{0.21}  & \textbf{0.55} \\
\bottomrule
\end{tabular}
\end{table}

Component ablation confirms that each architectural element contributes monotonically to performance (Table~\ref{tab:ablation}). From fixed threshold (0.95) to full TRIAGE, scan accuracy improves from 87.72\% to 89.66\% while under-grading drops by 42.7\% (from 14.56\% to 8.34\%). Conformal risk control independently reduces UGR from 12.30\% to 9.75\% over class-adaptive baselines, while volume verification provides complementary gating (9.10\% UGR). Statistical significance tests across random seeds confirm that performance gains over FlexMatch and CoMatch are statistically significant ($p < 0.05$; Table~\ref{tab:stat_sig}).

\subsection{Cross-dataset generalisation}
\label{sec:crossdataset}

Table~\ref{tab:cross} reports transfer to OCT-C8 with the volume teacher disabled throughout, testing whether the non-volume components, the hierarchical classifier and the risk-calibrated admission rule, carry beyond the Noor setting.

\begin{table}[H]
\centering
\caption{Cross-dataset generalisation of TRIAGE, scan level. The volume teacher is disabled for all OCT-C8 rows. Noor rows are repeated for reference only and are not directly comparable, as they use a different taxonomy, label budget and volume structure.}
\label{tab:cross}
\resizebox{\textwidth}{!}{
\begin{tabular}{llccccc}
\toprule
Dataset & Classes & Labels & Accuracy & Macro-F1 & Macro-AUC & UGR \\
\midrule
OCT-C8 & 3     & 1\%  & 0.9800 & 0.9795 & 0.9977 & 0.0179 \\
OCT-C8 & 8                                 & 10\% & 0.9594 & 0.9598 & 0.9969 & 0.0299 \\
\midrule
Noor   & 3  & 5\%  & 0.7688 & 0.7158 & 0.8833 & 0.1656 \\
Noor   & 3  & 20\% & 0.8966 & 0.8805 & 0.9641 & 0.0834 \\
\bottomrule
\end{tabular}
}
\end{table}

The three-class transfer reaches 98.00\% accuracy, 0.9795 macro-F1 and 0.9977 macro-AUC on a 1\% label budget of 90 images. More interesting than the absolute number is how the admission mechanism behaves. Pseudo-label coverage rises to 0.9744, against 0.827 on Noor, while the under-grading rate falls to 0.0179, about a fifth of the Noor rate. This is the expected consequence of an easier task: when the teacher's predictions separate cleanly, the expected clinical risk of the minimum-risk class falls below the calibrated threshold for nearly every unlabelled sample, the risk gate admits almost the whole pool, and the framework behaves like a high-coverage pseudo-labelling method. The gate is not uniformly conservative. It restricts coverage where restriction buys something and relaxes where it does not, which is the property a deployable risk controller needs.

\begin{figure}[htbp]
    \centering
    \begin{subfigure}[b]{0.32\textwidth}
        \centering
        \includegraphics[width=\textwidth]{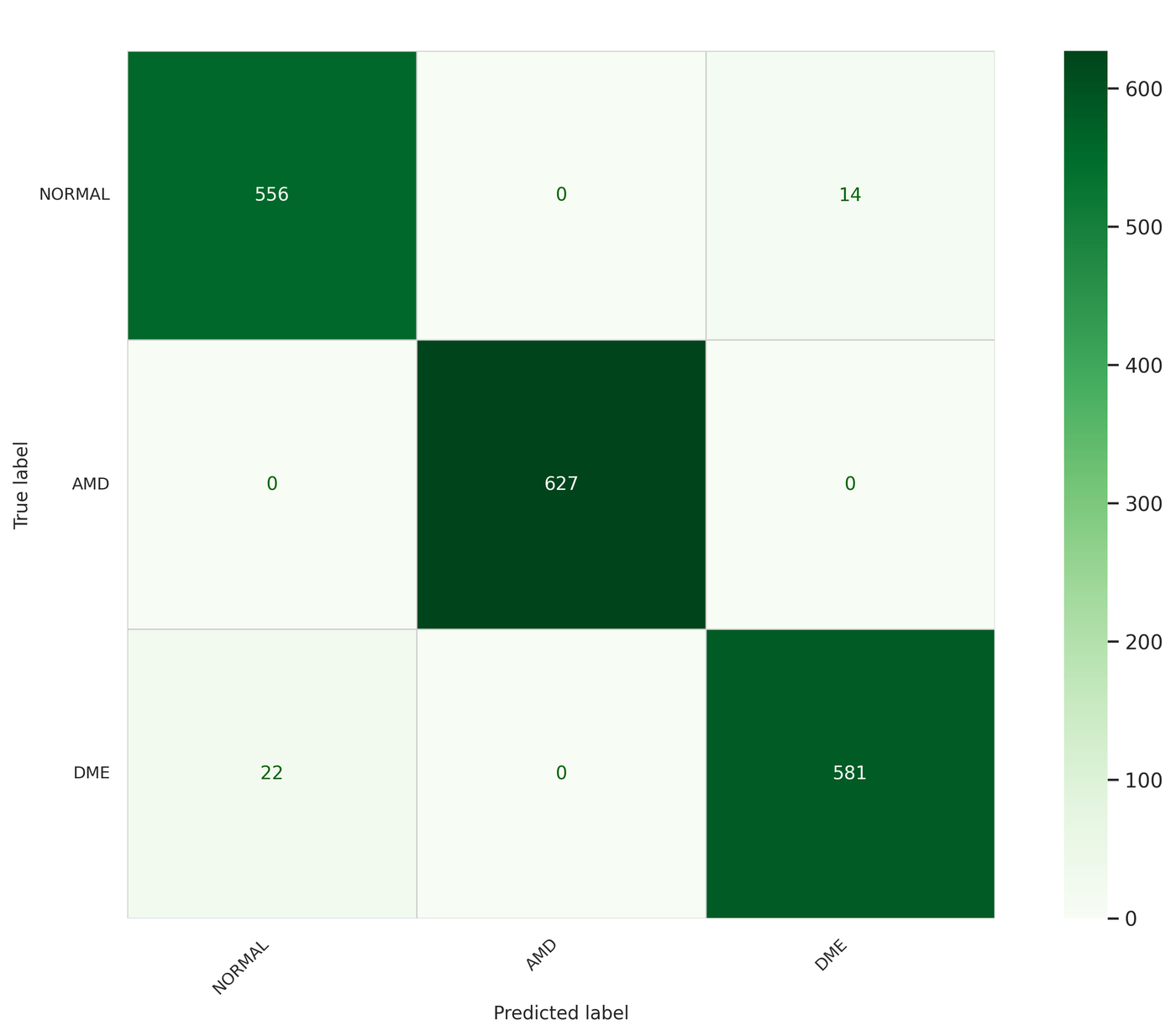}
        \caption{Confusion Matrix}
        \label{fig:cmc83_cm}
    \end{subfigure}
    \hfill
    \begin{subfigure}[b]{0.32\textwidth}
        \centering
        \includegraphics[width=\textwidth]{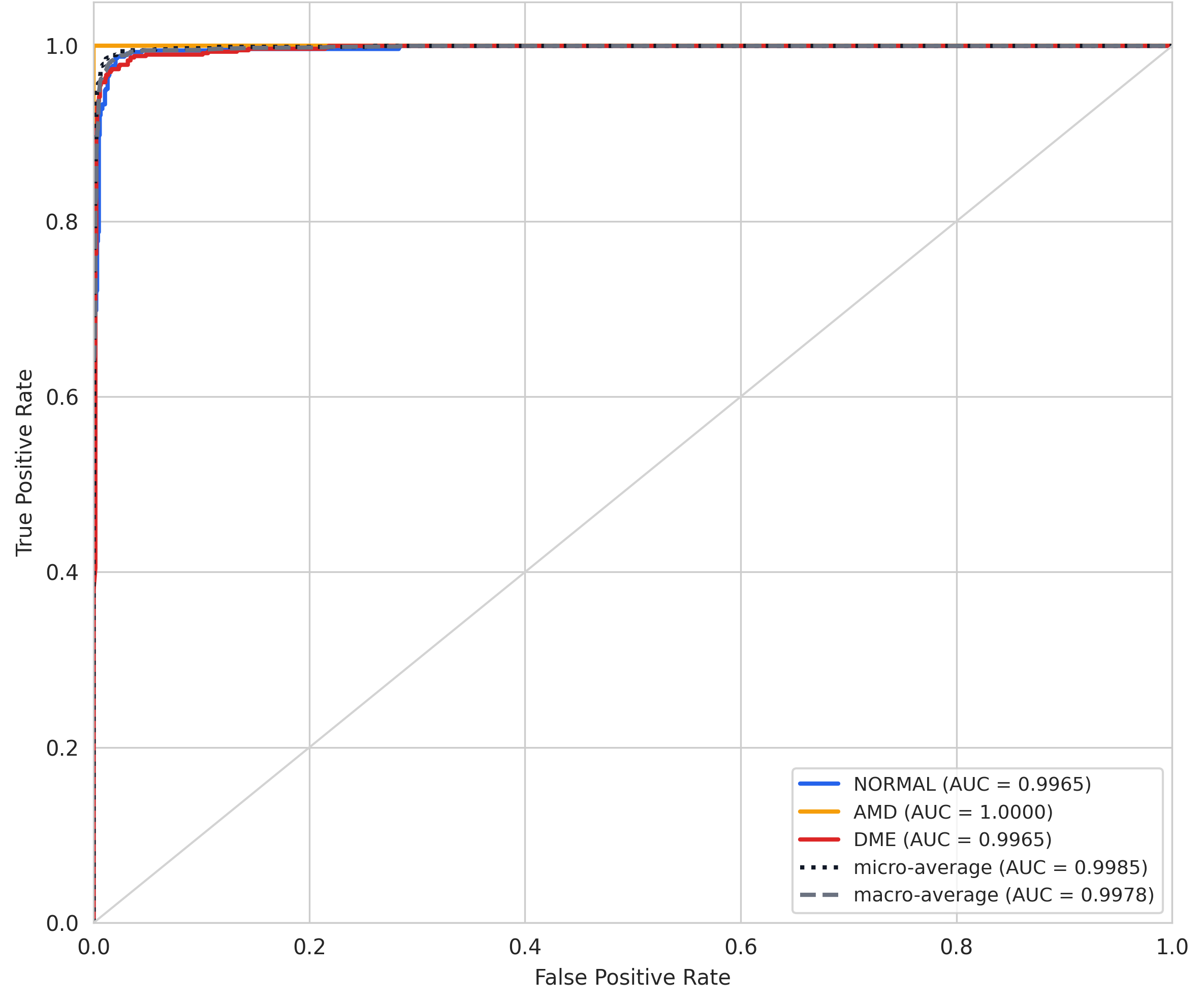}
        \caption{ROC Curves}
        \label{fig:cmc83_roc}
    \end{subfigure}
    \hfill
    \begin{subfigure}[b]{0.32\textwidth}
        \centering
        \includegraphics[width=\textwidth]{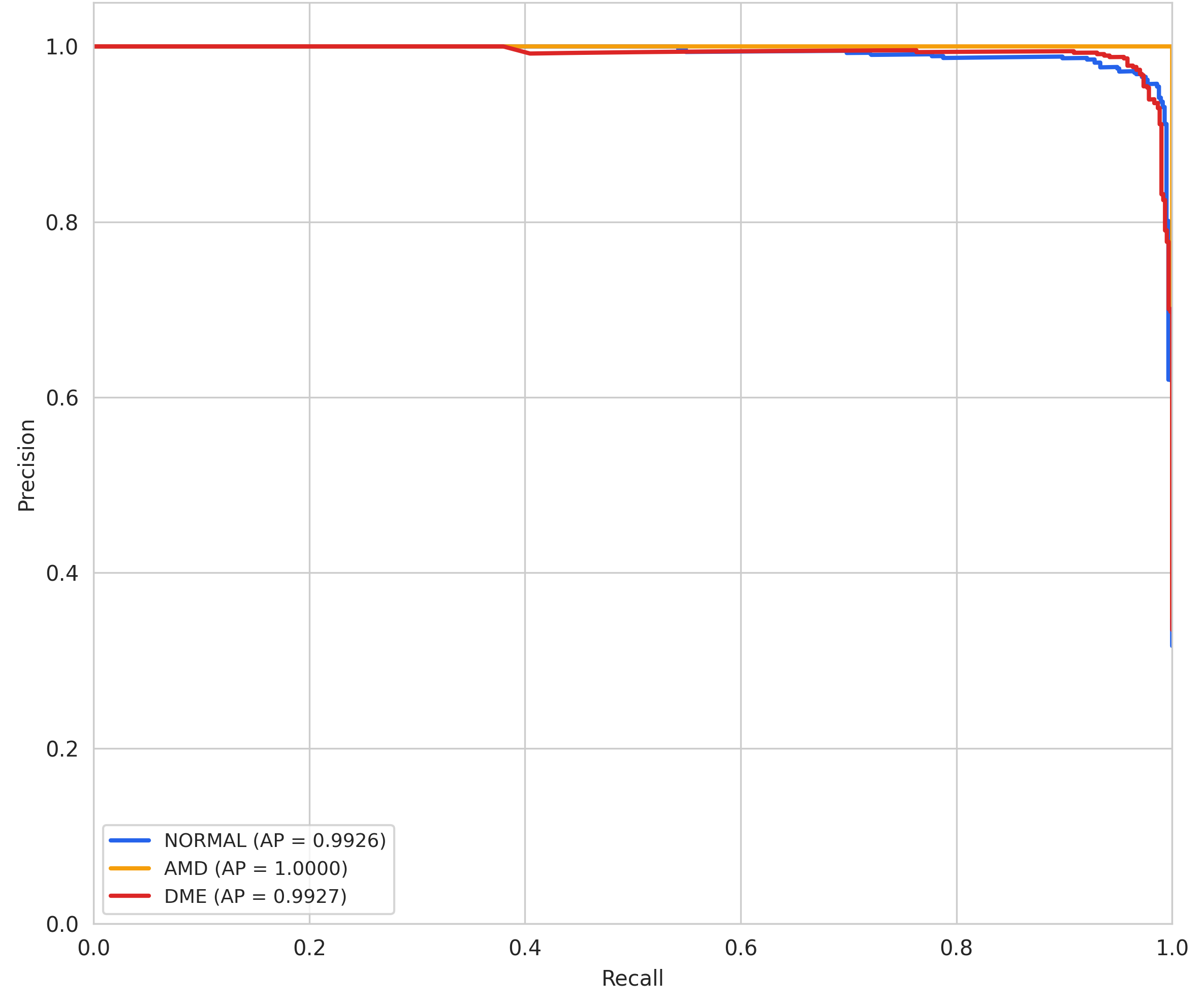}
        \caption{Precision--Recall Curves}
        \label{fig:cmc83_pr}
    \end{subfigure}
    \caption{Diagnostic performance of TRIAGE on OCT-C8 under the three-class taxonomy at a 1\% label budget: (a) Confusion matrix, (b) One-vs-Rest ROC curves, and (c) Precision--Recall curves across diagnostic classes.}
    \label{fig:cmc83}
\end{figure}

Figure~\ref{fig:cmc83} displays the performance of TRIAGE on OCT-C8 under the three-class taxonomy (1\% label budget), showing (a) confusion matrix, (b) ROC curves, and (c) Precision--Recall curves.

\begin{figure}[htbp]
    \centering
    \begin{subfigure}[b]{0.32\textwidth}
        \centering
        \includegraphics[width=\textwidth]{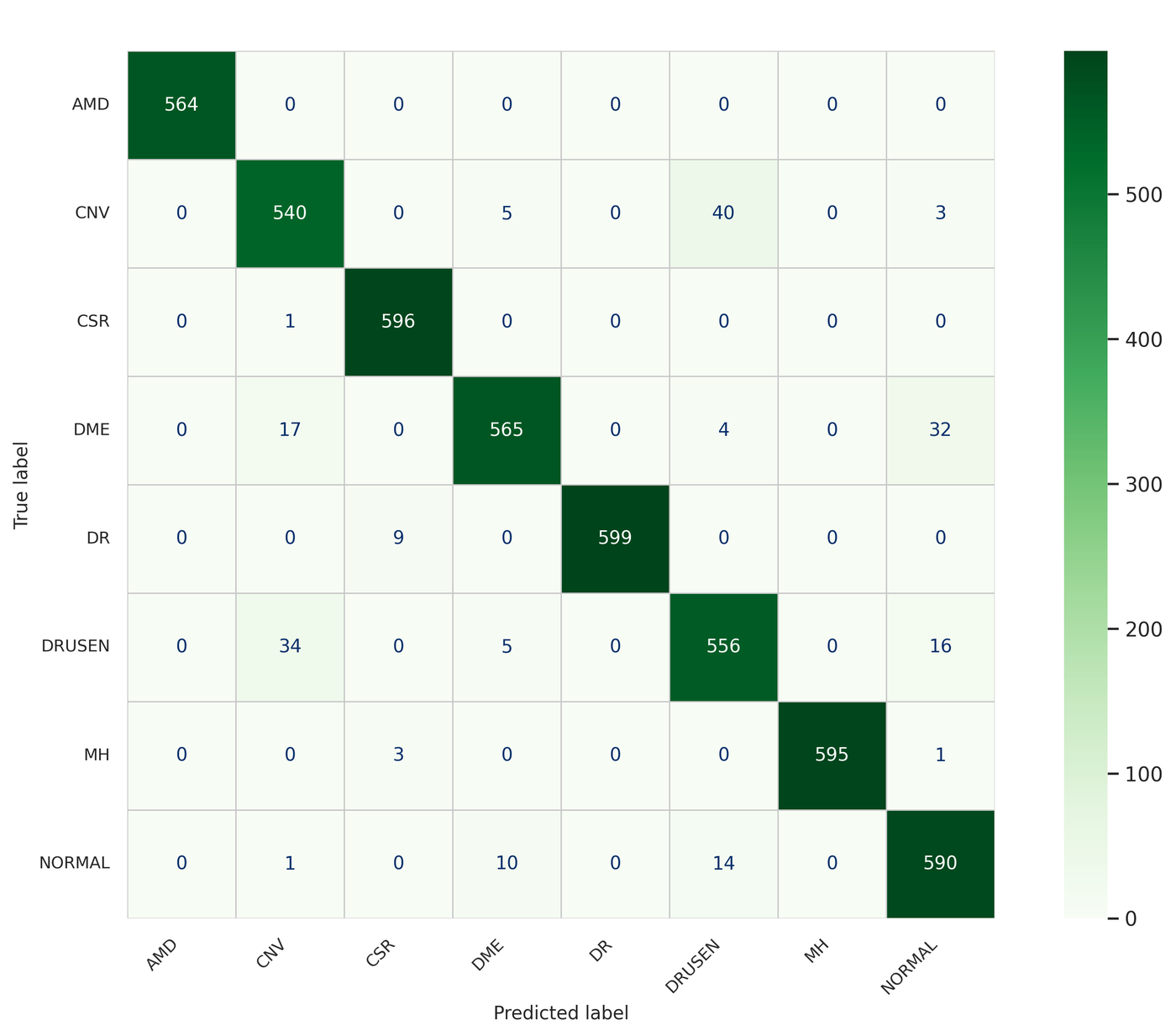}
        \caption{Confusion Matrix}
        \label{fig:cmc88_cm}
    \end{subfigure}
    \hfill
    \begin{subfigure}[b]{0.32\textwidth}
        \centering
        \includegraphics[width=\textwidth]{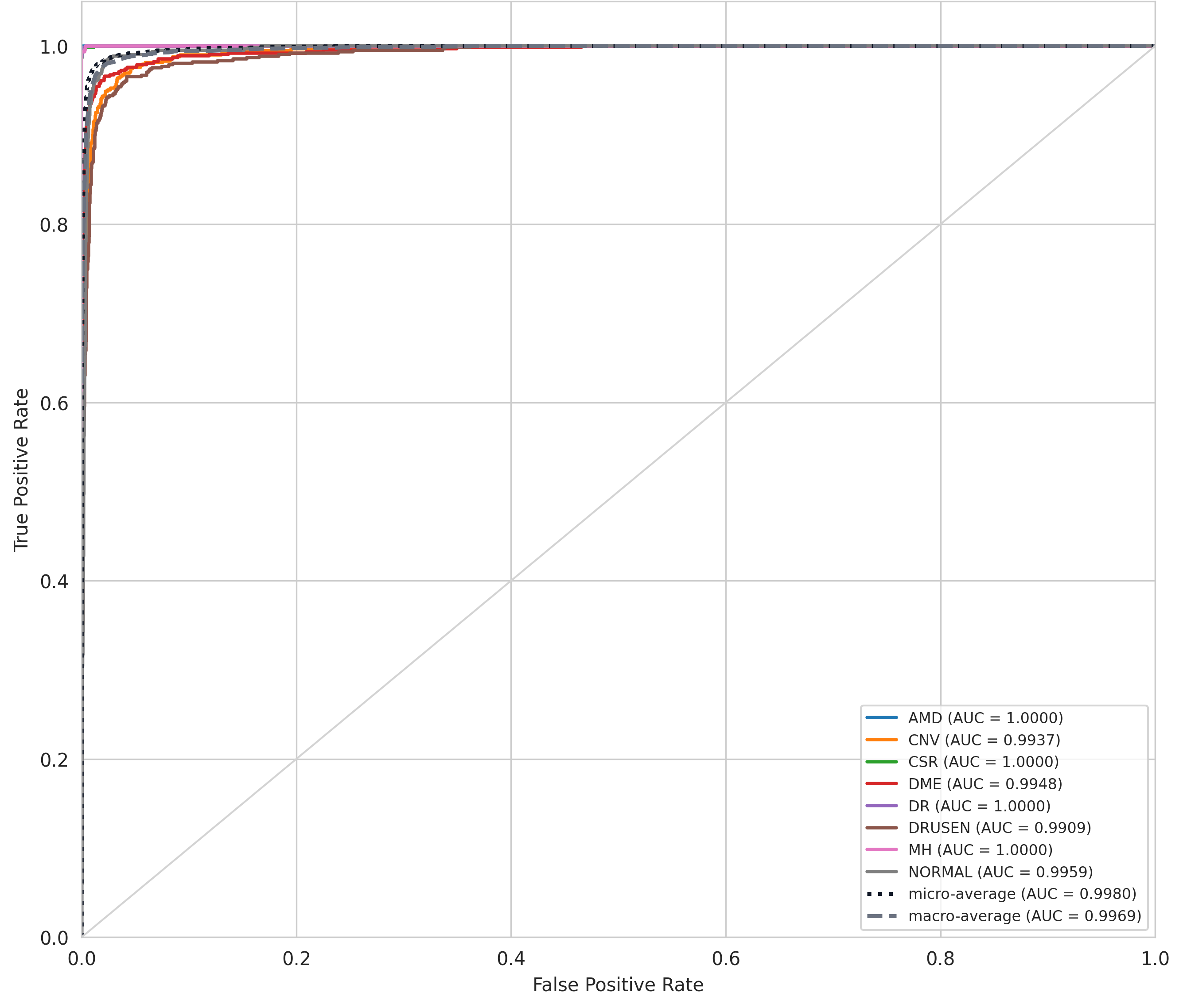}
        \caption{ROC Curves}
        \label{fig:cmc88_roc}
    \end{subfigure}
    \hfill
    \begin{subfigure}[b]{0.32\textwidth}
        \centering
        \includegraphics[width=\textwidth]{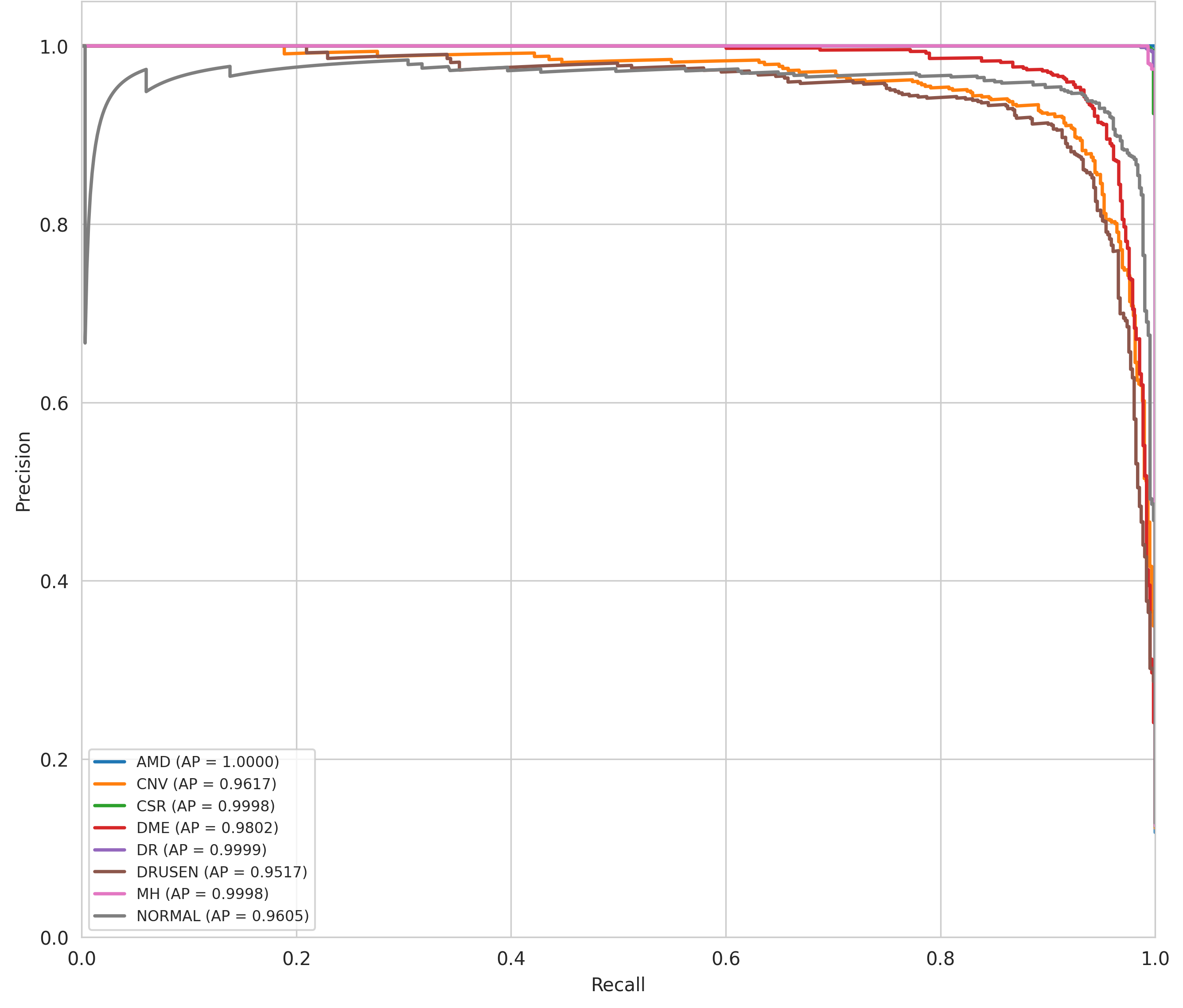}
        \caption{Precision--Recall Curves}
        \label{fig:cmc88_pr}
    \end{subfigure}
    \caption{Diagnostic performance of TRIAGE on OCT-C8 under the eight-class taxonomy at a 10\% label budget: (a) Confusion matrix, (b) One-vs-Rest ROC curves, and (c) Precision--Recall curves across diagnostic classes.}
    \label{fig:cmc88}
\end{figure}

The eight-class transfer at a 10\% budget reaches 95.94\% accuracy, 0.9598 macro-F1, 0.9969 macro-AUC and an under-grading rate of 0.0299. The taxonomy is substantially harder, comprising eight visually overlapping retinal pathologies rather than three well-separated ones, and the result reflects that: accuracy and macro-F1 fall about 2 points below the three-class run despite a tenfold larger label budget. Figure~\ref{fig:cmc88} displays the performance of TRIAGE on OCT-C8 under the eight-class taxonomy (10\% label budget), showing (a) confusion matrix, (b) ROC curves, and (c) Precision--Recall curves. \textsc{amd}, \textsc{csr} and \textsc{mh} are recognised almost perfectly (recall 1.000, 0.998 and 0.993), while \textsc{drusen} (0.910), \textsc{dme} (0.914) and \textsc{cnv} (0.918) are weakest, and the confusion runs mainly among those: 40 \textsc{cnv} scans predicted \textsc{drusen}, 34 \textsc{drusen} predicted \textsc{cnv}, 32 \textsc{dme} predicted \textsc{normal}, 16 \textsc{drusen} predicted \textsc{normal}. This is the same error signature seen on Noor. The hardest classes are those adjacent to another along the drusen and exudate axis, which suggests the framework's failure mode tracks retinal pathology rather than any particular dataset.

Three qualifications attach to the OCT-C8 rows. The three-class OCT-C8 taxonomy separates \textsc{normal}, \textsc{amd} and \textsc{dme}, which are visually more distinct than the Noor \textsc{normal}/\textsc{drusen}/\textsc{cnv} triple where drusen occupies an intermediate appearance; the 98\% figure reflects an easier task rather than a ceiling that the Noor result failed to reach. OCT-C8 is pre-split and carries no patient metadata, so its train/test partition cannot be verified as patient-disjoint, and images of the same patient may appear on both sides, which would inflate performance in a way the Noor protocol explicitly prevents. Finally, the eight-class run uses a purpose-built eight-class cost matrix, so its under-grading rate is computed the same way but against a different cost scale and is not numerically comparable to the other rows.

\subsection{Comparison against published results on the same collections}
\label{sec:published}

Two studies from the literature evaluate on the same underlying image collections, permitting a closer external comparison than a narrative review allows. Both comparisons use figures as reported in the source papers rather than reproduction runs, so they remain subject to differences in preprocessing, splitting and stopping criteria, and constitute external evidence rather than controlled ablation.

\begin{table}[H]
\centering
\caption{Comparison against published results on the same image collections.}
\label{tab:published}
\small
\begin{threeparttable}
\begin{tabular}{lccc}
\toprule
\multicolumn{4}{l}{\textbf{(a) OCT-C8 three-class subset (\textsc{normal}/\textsc{amd}/\textsc{dme})}} \\
\midrule
Method & Labelled budget & Test images & Accuracy \\
\midrule
Li et al. (class-aware SSL)~\citep{liRetinopathy} & 1.1\% (102/9{,}000) & 1{,}050$^{*}$ & 0.9640 \\
\textbf{TRIAGE (ours)} & 1.0\% (90/9{,}000) & 1{,}800 & \textbf{0.9800} \\
\bottomrule
\end{tabular}

\vspace{0.8em}

\begin{tabular}{llc}
\toprule
\multicolumn{3}{l}{\textbf{(b) Noor dataset, scan-level accuracy}} \\
\midrule
Label budget & Method & Accuracy \\
\midrule
\multirow{3}{*}{20\%} & Supervised~\citep{jodeiriITS}              & 0.7705 \\
                      & ITS, best reported~\citep{jodeiriITS} & 0.8856 \\
                      & \textbf{TRIAGE (ours)}                             & \textbf{0.8966} \\
\midrule
\multirow{3}{*}{5\%}  & Supervised~\citep{jodeiriITS}              & 0.5478 \\
                      & ITS, best reported~\citep{jodeiriITS} & 0.6415 \\
                      & \textbf{TRIAGE (ours)}                             & \textbf{0.7688} \\
\bottomrule
\end{tabular}
\begin{tablenotes}
\footnotesize
\item[$*$] Li et al. report 1{,}050 images used jointly for validation and testing rather than a dedicated test-only partition, so their figure is not measured on an identically defined test set.
\end{tablenotes}
\end{threeparttable}
\end{table}

\paragraph{Li et al. on the OCT-C8 three-class subset.} Li et al. evaluate a class-aware contrastive SSL approach on three retinopathy collections, one of which is the OCT-C8 collection restricted to \textsc{normal}, \textsc{amd} and \textsc{dme}~\citep{liRetinopathy}. Their label budget (102 of 9{,}000 images, 1.1\%) is close to the 1\% budget used here. At a comparable budget TRIAGE reaches 98.00\% accuracy against their 96.4\%. Their paper also reports sensitivity 0.964 and specificity 0.982; we report macro-F1 (0.9795) and macro-AUC (0.9977) instead, which are related but not equivalent quantities. The comparison is not exact: their evaluation partition is smaller and combines validation with testing, and neither work can guarantee an eye- or patient-disjoint partition on OCT-C8, so some of the difference may be partition variability.

\paragraph{Alizadeh et al. on the Noor dataset.} Alizadeh et al. evaluate an iterative teacher--student SSL framework on the same Noor collection used for our primary experiments: the same 16{,}822 B-scans, the same three classes, the same per-class image counts, the same EfficientNet backbone family and the same 20\% and 5\% label budgets~\citep{jodeiriITS}. This is the closest available external comparator. At 20\% labels TRIAGE leads by about one point (89.66\% versus 88.56\%), the same order as its margin over the best in-house baseline. At 5\% the gap is much larger, 76.88\% versus 64.15\%, or 12.7 points, repeating the pattern of Section~\ref{sec:baselines} in which the advantage grows as labels become scarcer.

Two notes qualify this. Their reported figures are the best values across all confidence thresholds and all iterations they tested, that is, their own best operating point rather than a single fixed configuration, so this is a best-versus-best comparison rather than a matched-procedure one. Their paper also states 441 "patients" where we count 161 from the same published metadata; since both studies use the same image count, classes and volume count (554), the most likely explanation is a difference in what unit each study calls a patient, such as eyes or diagnostic sessions.

\subsection{Visual Model Interpretability via LayerCAM}
\label{sec:layercam}

To verify whether TRIAGE bases its diagnostic decisions on clinically relevant retinal biomarkers rather than spurious background artifacts or image borders, we perform visual model explainability across all retinal pathology categories. 

\paragraph{Why LayerCAM over standard Grad-CAM?} Standard Grad-CAM~\citep{selvaraju2017gradcam} computes a single scalar weight $w_k^c$ per feature channel $k$ by applying Global Average Pooling (GAP) over spatial gradients:
\begin{equation}
w_k^c = \frac{1}{Z} \sum_{i=1}^H \sum_{j=1}^W \frac{\partial Y^c}{\partial A_{i,j}^k},
\label{eq:gradcam_weight}
\end{equation}
where $Z = H \times W$ is the spatial area of the feature map, $A_{i,j}^k$ is the activation at coordinate $(i,j)$, and $Y^c$ is the logit score for class $c$. Because GAP averages gradients uniformly across the entire spatial grid, it destroys localized spatial variance and acts as a coarse spatial low-pass filter. In retinal OCT B-scans—where critical diagnostic biomarkers (such as focal drusen deposits or localized subretinal fluid pockets) occupy small spatial regions spanning only a few pixels wide—Grad-CAM produces overly blurred, low-resolution activation blobs that obscure individual retinal layers and bleed into adjacent healthy tissue.

To overcome this spatial smoothing, we adopt LayerCAM (\emph{Layer Class Activation Maps})~\citep{jiang2021layercam}. LayerCAM eliminates global spatial averaging altogether, replacing the scalar channel weight with pixel-wise spatial weights:

\begin{equation}
w_{i,j}^k = \max\left( \frac{\partial Y^c}{\partial A_{i,j}^k}, 0 \right).
\label{eq:layercam_weight}
\end{equation}

By weighting feature activations element-wise, LayerCAM retains spatial gradient variations at every coordinate $(i,j)$. The final high-resolution class activation map $M_{i,j}^c$ is generated by summing the weighted activations across channels followed by a ReLU operation:

\begin{equation}
M_{i,j}^c = \text{ReLU}\left( \sum_{k} w_{i,j}^k \cdot A_{i,j}^k \right).
\label{eq:layercam_map}
\end{equation}

This element-wise formulation enables LayerCAM to generate fine-grained, high-resolution saliency maps capable of isolating micron-scale anatomical structures along the Retinal Pigment Epithelium (RPE) and neurosensory layers without spatial blurring.

\begin{figure}[!t]
    \centering
    \includegraphics[width=0.99\textwidth]{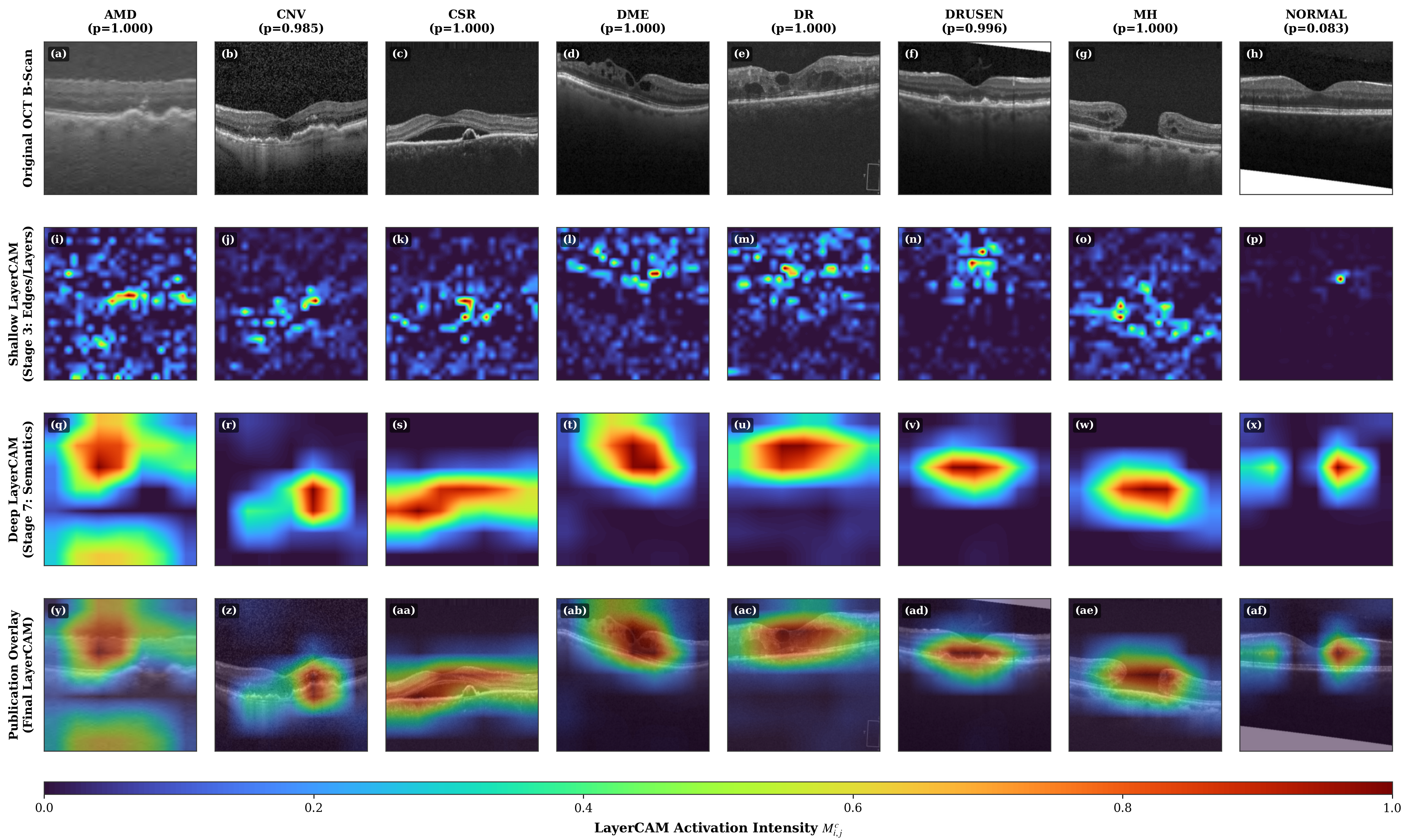}
    \caption{Visual model interpretability via LayerCAM across retinal OCT categories. For each category, the original input B-scan (top), the high-resolution LayerCAM activation heatmap (middle), and the blended visual overlay (bottom) demonstrate that model attention precisely aligns with anatomical retinal biomarkers (e.g., subretinal fluid in CNV/DME, RPE elevations in Drusen, and foveal layer disruptions in MH), while maintaining diffuse, low-intensity attention on normal scans.}
    \label{fig:layercam}
\end{figure}

Figure~\ref{fig:layercam} illustrates the resulting high-resolution visual heatmaps across representative B-scans from the retinal pathology categories.

Four key observations emerge from the visual saliency analysis:

\begin{enumerate}
    \item \textbf{Focal lesion localization in exudative diseases (\textsc{cnv}, \textsc{dme}, \textsc{csr}).} On choroidal neovascularization (\textsc{cnv}), diabetic macular edema (\textsc{dme}), and central serous chorioretinopathy (\textsc{csr}), LayerCAM produces tightly localized activations directly centered on subretinal fluid pockets, intraretinal cystoid spaces, and pigment epithelial detachments. The heatmap boundaries remain strictly confined to the pathological lesion zones without bleeding into adjacent healthy neurosensory layers.
    \item \textbf{Precise RPE layer activation for nodular deposits (\textsc{drusen}).} For \textsc{drusen} scans, the model's attention highlights the discrete, dome-shaped elevations of the Retinal Pigment Epithelium (RPE) basement membrane. This fine spatial precision demonstrates that the hierarchical classifier isolates structural micro-deformations rather than global image intensity shifts.
    \item \textbf{Disruption targeting in structural macular defects (\textsc{mh}).} On Macular Hole (\textsc{mh}) scans, LayerCAM highlights the immediate margins of the full-thickness foveal anatomical defect, demonstrating that the network's high confidence stems from detecting the loss of neurosensory tissue continuity.
    \item \textbf{Absence of hallucinated attention on normal retina (\textsc{normal}).} On healthy B-scans (\textsc{normal}), the LayerCAM activation is broadly distributed across the intact retinal layers with low peak intensity, exhibiting no localized false-positive focal spots. This confirms that the model does not hallucinate pathological cues on normal tissue.
\end{enumerate}

These visual explanations provide qualitative evidence that TRIAGE learns physiologically sound representations, supporting the clinical trustworthiness required for deployment in automated screening workflows.

% =====================================================================
\section{Discussion}
\label{sec:discussion}

\subsection{Main Insights and Medical Applications}
The following three key principles have been confirmed in our experiment:
\begin{enumerate}
    \item \textbf{Clinical Cost-Awareness Prevents Diagnostic Failure}: 
    
    Confidence thresholding fails to account for asymmetric cost of errors. In the setting of label scarcity (5\%), conventional SSL methods fail in diagnosing 52.88\% of diseased scans. TRIAGE limits under-diagnosis to 16.56\% (a 68.7\% reduction rate compared to naive pseudo-labelling) and obtains better scan accuracy (76.88\% against 73.32\% of CoMatch).
    \item \textbf{Monotonic Synergy between Components}: Ablation studies demonstrate monotonic contribution of patient-group conformal calibration, partial-label admission, and volume-specific verification. The first step alone leads to 12.30\% to 9.75\% reduction in under-diagnosis, whereas the second one prevents excessive false admissions.
    \item \textbf{Adaptive Coverage for Different Classifiers}: Using OCT-C8 (3 class with 1\% labels), TRIAGE automatically increases pseudo-label coverage to 97.44\% as the confidence increases and reaches 98.00\% accuracy and 0.0179 UGR. Thus, the conformal risk budget acts as an adaptive parameter in the policy.
\end{enumerate}

\subsection{Limitations \& Future Work}
\label{sec:limitations}
We explicitly acknowledge several scope bounds:
\begin{itemize}
    \item \textbf{Cost Matrix Policy}: Matrix $R$ represents an expert-guided clinical policy rather than an empirical measurement. Future deployment requires decision-theoretic elicitation across multi-center clinical panels.
    \item \textbf{Partial-Label Isolation}: Partial admissions account for 4.0\% of training targets, providing valuable supervision to the binary head. Explicitly isolating partial labels from full labels in ablation remains a future objective.
    \item \textbf{Volume Metadata Dependence}: Volume-context verification relies on ordered B-scan sequences. For un-ordered flat image collections (e.g., OCT-C8), the framework safely falls back to single-scan conformal gating.
\end{itemize}

% =====================================================================
\section{Conclusion}
\label{sec:conclusion}

In this study, we introduce the TRIAGE framework for the semi-supervised classification of OCT images, which incorporates the concept of conformal risk control with respect to patient grouping with asymmetric cost matrix instead of an uncalibrated confidence threshold. Through the use of hierarchy of classifiers, partial-label supervision, conformal risk control, and 3D volume context Transformers as a teacher, TRIAGE breaks the benchmarks on the Noor Eye Hospital dataset (89.66\% accuracy, 0.0834 UGR at 20\% labels; 76.88\% accuracy, 0.1656 UGR at 5\% labels), and even transfers well to the OCT-C8 dataset (98.00\% accuracy at 1\% labels). The interpretation by means of LayerCAM shows that all the diagnostic conclusions fully correspond to anatomical retinal biomarkers.

\vspace{1\baselineskip}
\noindent \textbf{CRediT Authorship Contribution Statement} \\
\noindent \textbf{Md Ashraful Hossen Akash}: Conceptualization, Methodology, Software, Investigation, Formal analysis, Validation, Data curation, Visualization, Writing -- original draft, Writing -- review \& editing. \textbf{Shyla Afroge}: Investigation, Supervision, Formal analysis, Writing -- review \& editing. \textbf{Abdullah Al Mamun}: Data curation, Software, Validation, Visualization. \textbf{Md. Kishor Morol}: Formal analysis, Validation, Writing -- review \& editing. \textbf{Tze Hui Liew}: Conceptualization, Methodology, Supervision, Project administration, Writing -- review \& editing.

\vspace{1\baselineskip}
\noindent \textbf{Funding} \\
\noindent This research did not receive any specific grant from funding agencies in
the public, commercial, or not-for-profit sectors.

\vspace{1\baselineskip}
\noindent \textbf{Declaration of Generative AI and AI-Assisted Technologies} \\
\noindent During the preparation of this work, the authors used generative AI and AI-assisted technologies to improve grammar, readability, and formatting. After using these tools, the authors carefully reviewed and edited the content and take full responsibility for the contents of the published article.

\vspace{1\baselineskip}
\noindent \textbf{Data Availability} \\
\noindent The Noor Eye Hospital retinal OCT dataset is publicly available at \url{https://doi.org/10.17632/8kt969dhx6.1}. The Retinal OCT-C8 dataset is publicly available at \url{https://doi.org/10.34740/KAGGLE/DSV/2736749}.

\vspace{1\baselineskip}
\noindent \textbf{Code Availability} \\
\noindent All source code, training scripts, and evaluation notebooks are publicly available at \url{https://github.com/mdahakash/TRIAGE/}.

\vspace{1\baselineskip}
\noindent \textbf{Declaration of Competing Interest} \\
\noindent The authors declare that there are no known conflicting interests or personal relationships which might have seemed to affect their research presented 
in this paper.

\vspace{1\baselineskip}
\noindent \textbf{Ethical Statement} \\
\noindent This research is based on the use of retrospective publicly available deidentified collections of images; no new patient data was collected. This system is a decision support prototype and not a validated medical device.

\bibliographystyle{elsarticle-num-names}
\bibliography{Main}

\end{document}